\documentclass[10pt,twocolumn,letterpaper]{article}

\usepackage[algorithms]{wacv}

\usepackage{array}
\usepackage{multirow}
\usepackage{amsfonts}
\usepackage[ruled,linesnumbered]{algorithm2e}
\usepackage{makecell}
\usepackage{textcomp}
\usepackage{stfloats}
\usepackage{cuted}
\usepackage{placeins}
\usepackage{verbatim}
\usepackage{mathrsfs}
\usepackage{tikz}
\usepackage{epstopdf}
\usepackage{arydshln}	
\usepackage{bm}
\usepackage{nicematrix}
\usetikzlibrary{calc, positioning, fit, arrows.meta}

\definecolor{wacvblue}{rgb}{0.21,0.49,0.74}
\usepackage[breaklinks,colorlinks,allcolors=wacvblue]{hyperref}

\def\wacvPaperID{1708}
\def\confName{WACV}
\def\confYear{2027}

\title{Preserve, Then Resolve: Many-to-Many Association and \\ Robust Estimation with General-Purpose Visual Features}
\author{%
	Haodong JIANG \qquad Mingzhe LI \qquad Junfeng WU\\
	School of Data Science, CUHK~(SZ)\\
	\texttt{\{haodongjiang, mingzheli2\}@link.cuhk.edu.cn, junfengwu@cuhk.edu.cn}}
\hypersetup{
	pdfauthor={Haodong Jiang, Mingzhe Li, Junfeng Wu},
	pdftitle={Preserve, Then Resolve: Many-to-Many Association and Robust Estimation with General-Purpose Visual Features}
}

\newtheorem{assumption}{Assumption}

\newcommand\bfR{\mathbf{R}}

\newcommand\bfone{\mathbf{1}}

\newcommand\bft{\mathbf{t}}

\newcommand\bfx{\mathbf{x}}
\newcommand\bfy{\mathbf{y}}

\newcommand{\bfth}{\boldsymbol{\theta}}

\definecolor{LightBlue}{rgb}{0.1,0.5,1}
\definecolor{LightGreen}{HTML}{5a9678}
\definecolor{LightYellow}{HTML}{fbc02d}
\definecolor{LightPurple}{HTML}{9664B4}

\definecolor{RadarBlue}{HTML}{2A78D6}
\definecolor{RadarAqua}{HTML}{1BAF7A}
\definecolor{RadarYellow}{HTML}{EDA100}
\definecolor{RadarGreen}{HTML}{008300}

\newcommand{\wacvteaser}{%
	\par\vspace{-0.5em}%
	\begin{minipage}{\textwidth}
		\centering
		\captionsetup{type=figure}
		\begin{subfigure}[b]{0.54\linewidth}
			\centering
			\includegraphics[width=\linewidth]{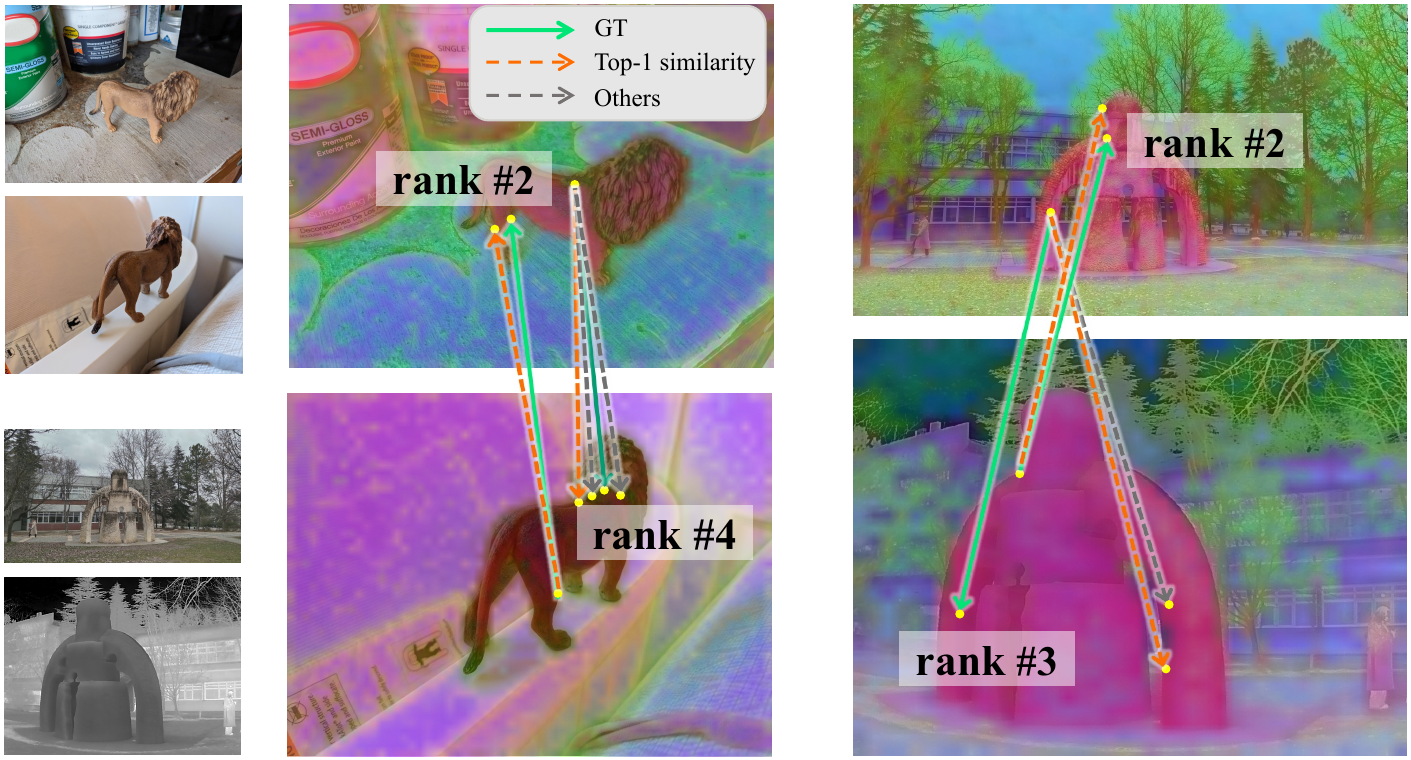}
			\caption{}
			\label{fig::teaser_examples}
		\end{subfigure}
		\hfill
		\begin{subfigure}[b]{0.45\linewidth}
			\centering
			\includegraphics[width=\linewidth]{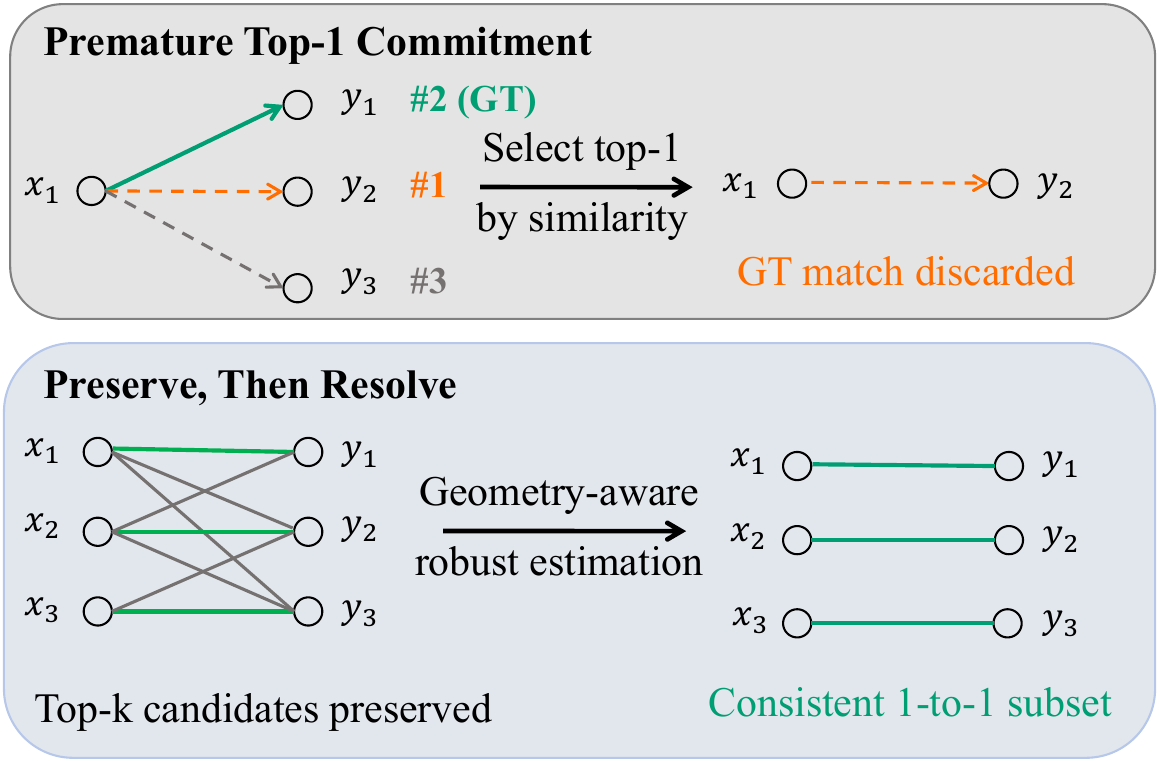}
			\caption{}
			\label{fig::teaser_principle}
		\end{subfigure}
		\addtocounter{figure}{-1}
		\captionof{figure}{(a) Bias-mitigated intermediate-layer DINOv3 features remain semantically consistent across
		background variation and visible--thermal modality shifts~(top three PCA components of DINOv3 patch features are shown as RGB). However, geometrically correct correspondences~(green solid lines) often fall below rank one in cosine similarity~(orange dash lines). (b)~A premature rank-one decision may therefore discard useful geometric evidence. We instead preserve multiple candidates in a many-to-many association graph and defer their disambiguation to robust estimation, which favors geometrically consistent support satisfying the 1-to-1 constraint.}
		\label{fig::teaser}
	\end{minipage}%
	\par\vspace{0.5em}%
}
\makeatletter
\apptocmd{\@maketitle}{\wacvteaser}{}{\PackageError{main_wacv}{Unable to attach teaser to title block}{}}
\makeatother

\begin{document}

	\maketitle

	\begin{abstract}
	The semantic transferability of general-purpose visual features does not guarantee geometric consistency across images. Using frozen DINOv3 features, we show that geometrically correct correspondences often fall below rank one in cosine similarity yet remain within a small top-$K$ candidate set. This motivates a preserve-then-resolve design: we retain multiple candidates in a many-to-many~(m-to-m) association graph and defer their disambiguation to robust estimation. We study m-to-m robust estimation from a probabilistic perspective. We interpret the existing Matching Cardinality Maximization~(MCM) mechanism as a dominant-cardinality approximation to likelihood maximization and propose a faster, real-valued mechanism called Harmonic Consensus Maximization~(HCM). A two-stage LO-RANSAC uses HCM for candidate sourcing and MCM for graph-aware selection. We evaluate end-task gains in relative-pose estimation, where our pipeline consistently improves rank-one baselines for DINOv2, DINOv3~\cite{oquab2023dinov2,simeoni2025dinov3}, V-JEPA~2.1~\cite{murlabadia2026vjepa2_1}, and SigLIP~2~\cite{tschannen2025siglip2}. Code is available at \url{https://github.com/LIAS-CUHKSZ/preserve_then_resolve}.
	\end{abstract}
	
	\section{Introduction}
	\begin{flushright}
		\begin{minipage}{0.96\columnwidth}
			\small\itshape
			``Who can wait quietly while the mud settles?\\
			Who can remain still until the moment of action?''\par
			\vspace{0.15em}
			\makebox[\linewidth][r]{\normalfont\scriptsize ---\,Lao Tzu, \textit{Tao Te Ching}, ch.~15 (trans. Gia-fu Feng and Jane English)}
		\end{minipage}
	\end{flushright}
	Visual foundation models~\cite{radford2021learning,oquab2023dinov2,simeoni2025dinov3,tschannen2025siglip2,murlabadia2026vjepa2_1} trained on large and diverse image collections have demonstrated strong zero-shot transfer in image retrieval, semantic segmentation, and semantic correspondence~\cite{amir2021deep,tang2023emergent,zhang2023tale,keetha2023anyloc}. However, this impressive semantic transferability does not guarantee consistent 3D correspondence~\cite{el2024probing,zhang2024telling}. As shown in Fig.~\ref{fig::teaser}, different arms of a statue may be confused across viewpoints. This tension motivates our central question:
	
	\textit{How should data association and robust mechanisms be co-designed to better exploit semantically transferable yet geometrically ambiguous features in geometric estimation?}

	Following prior work~\cite{amir2021deep,zhang2023tale,el2024probing,zhang2024telling}, we investigate this question by deploying frozen general-purpose features, without task-specific feature adaptation, in a zero-shot image-matching pipeline. We use DINOv3~\cite{simeoni2025dinov3} as our primary diagnostic and examine whether the resulting downstream gains extend to other representations. On a dedicated diagnostic split sampled from NAVI-Wild~\cite{jampani2023navi}, we find that layer selection and positional-bias correction mitigate but do not eliminate geometric ambiguity: the real association often falls below rank one in cosine similarity while remaining within a small top-$K$ candidate set. This observation motivates a \textbf{preserve-then-resolve} design.

	\emph{Preserve ambiguity in association.} Rather than commit prematurely to a single similarity-ranked match, we retain multiple candidates in a many-to-many~(m-to-m) association graph. This design increases the chance that useful geometric evidence survives association, but also admits more spurious edges and defers disambiguation to estimation. Similar m-to-m associations have previously arisen from unknown correspondences~\cite{pathak2010fast,campbell2018globally,wang2022certifiably,yang2020teaser}, descriptor quantization~\cite{mcilroy2010deterministic,camposeco2019hybrid,jiang2025score}, and repetitive patterns~\cite{fredriksson2016optimal}.
	
	\emph{Resolve ambiguity in estimation.} Physical correspondences remain 1-to-1 even when candidate associtiations are m-to-m. MCM accounts for this structure by scoring each hypothesis with the cardinality of a maximum matching in its inlier subgraph. Our probabilistic formulation interprets the MCM score as a dominant-cardinality surrogate and motivates HCM, a lower-cost, real-valued score that can distinguish hypotheses with the same discrete MCM cardinality. We combine HCM and MCM in a two-stage LO-RANSAC algorithm for m-to-m robust estimation.

	Together, these two stages establish a deliberate division of labor: general-purpose features propose a small set of semantically plausible candidates, whereas graph-aware and likelihood-inspired robust mechanisms identify candidates that jointly support a physical model. We evaluate downstream gains with relative-pose estimation on six datasets, using controlled comparisons to examine candidate retention, ambiguity-aware estimation, and representation compatibility. Our contributions are summarized as follows:
	\begin{itemize}
		\item \textbf{Preserve-then-resolve design for general-purpose features.}
		Using frozen DINOv3, we identify persistent rank ambiguity: representation-side improvements enhance geometric correspondence but do not prevent valid matches from falling below rank one. We translate this observation into an association--estimation design that preserves plausible candidates in an m-to-m graph and defers their disambiguation to geometry.

		\item \textbf{Likelihood-based scoring for m-to-m estimation.}
		We develop a probabilistic formulation for robust estimation under m-to-m data association, interpret the discrete MCM score as a dominant-cardinality surrogate for the otherwise intractable log-likelihood, and derive HCM as a linear-time, real-valued scorer.
  
		\item \textbf{Efficient realization and controlled validation.}
		We combine efficient HCM exploration with graph-aware MCM selection in a two-stage LO-RANSAC. Controlled relative-pose experiments characterize the benefit and limits of candidate
		preservation, demonstrate the complementary roles of HCM and MCM, and show consistent improvements over representation-matched 1-to-1 baselines for four evaluated general-purpose representations.
	\end{itemize}
	
	\begin{figure*}[!htbp]
		\centering
		\includegraphics[width=0.9\textwidth]{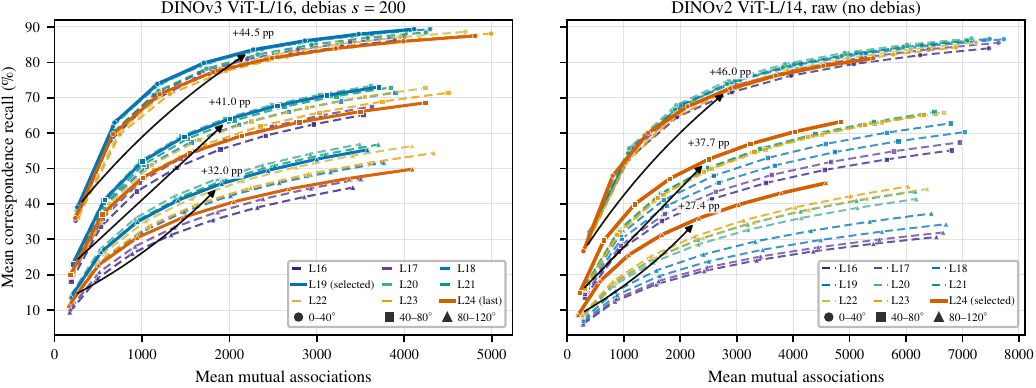}
		\caption{Correspondence-recall versus association-count trade-offs produced by the MKNN test using corrected DINOv3 and raw DINOv2 features from ViT-L layers~16 to 24. Each trajectory connects the operating points for $K=1,\ldots,8$. Circle, square, and triangle markers denote three view-point variation bins. Solid trajectories denote the selected and final layers for each model. Arrows connect the $K=1$ and $K=5$ points for the selected configurations; labels above the arrows report the absolute recall gains in percentage points~(pp).}
		\label{fig::layer_selection}
	\end{figure*}
	\section{Feature Association Ambiguity}\label{section::DINO_diagnosis}
	We adopt the DINOv3 ViT-L/16 model as our primary diagnostic and include DINOv2 ViT-L/14 with registers as a reference. Following~\cite{simeoni2025dinov3,el2024probing}, we use NAVI-Wild~\cite{jampani2023navi}, an object-centric benchmark with variations in viewpoint, background, and camera model. Our dedicated diagnostic split contains 1,500 image pairs, with 500 pairs in each of three camera-angular-distance bins: $[0^\circ, 40^\circ)$, $[40^\circ, 80^\circ)$, and $[80^\circ, 120^\circ)$, and image-disjoint from the estimtion split.
  
	We use the following evaluation protocol. \textbf{First}, we construct patch associations using the mutual $K$-nearest-neighbor~(MKNN) test: an association is retained only if each patch lies among the other patch's top-$K$ neighbors under cosine similarity. \textbf{Next}, we vary $K$ and record both the fraction of \textit{eligible patches} recovered by at least one geometrically correct association~(\textit{correspondence recall}) and the number of retained associations. A patch is eligible if its center has valid depth and the other view contains at least one valid patch within a 3D distance of $\tau$ cm. An eligible patch is considered recovered if it is incident to \emph{at least one} selected association satisfying this geometric threshold. We compute correspondence recall separately for $\tau\in\{1,2,5\}$ cm and report the mean across thresholds. \textit{Supplementary}~\ref{append::layer_selection} details the sampling, patch geometry, aggregation, and parameter sweep and compares MKNN with one-way and bidirectional-union KNN.

	\textbf{Representation-side improvements are helpful but insufficient.} We first select the ViT-L layer best suited to geometric correspondence and, for DINOv3, apply the training-free correction proposed by prior work~\cite{cuttano2026insid3} to mitigate its inherent positional bias. This correction is controlled by the subspace rank~$s$. Because DINOv2 has been reported to exhibit negligible positional bias~\cite{cuttano2026insid3}, we evaluate its features without this correction. Figure~\ref{fig::layer_selection} shows DINOv3 results at $s=200$ alongside raw DINOv2 features; \textit{Supplementary}~\ref{append::layer_selection} provides results for all evaluated subspace ranks. These results yield three observations. \textbf{First}, across the evaluated debiasing ranks, DINOv3 layer~19 remains close to the empirical upper-left recovery frontier, with its advantage over the final layer being largest under substantial viewpoint changes. For DINOv2, the final layer provides the strongest trade-off. \textbf{Second}, most of the gain from correcting DINOv3 features is already achieved at $s=100$; differences over $s=100$--$600$ are small and non-monotonic. We select $s=200$ from this stable range. \textbf{Third}, under the selected configurations, DINOv3 provides a stronger recall--association-count trade-off than DINOv2. Nevertheless, both models show decreasing correspondence recall as viewpoint separation increases, a failure pattern also reported for other general-purpose features~\cite{el2024probing}.

	\textbf{Geometrically valid candidates often lie beyond rank one.} For corrected layer-19 DINOv3 features~($s=200$), increasing $K$ from 1 to 5 raises correspondence recall by $44.5/41.0/32.0$ pp across the three viewpoint bins. These substantial gains indicate that geometrically valid counterparts often fail to rank first by feature similarity but remain within a small neighborhood of candidates. The DINOv2 curves exhibit the same recovery behavior, indicating that this is not unique to DINOv3. These results motivate preserving several competing associations instead of enforcing a rank-one decision. Preserving them improves the chance that geometrically valid evidence survives association, but it also produces a low-precision m-to-m association graph. We defer disambiguiation to tailored robust mechanisms.

	\section{M-to-M Robust Estimation}\label{section::theory}
	Consider estimating a parameter $\bfth \in \mathbb{R}^d$ from a set of data associations $(\mathbf{x}_i, \mathbf{y}_j)$, of which only an unknown subset is physically valid. Let $f_{\bfth}(\mathbf{x}_i, \mathbf{y}_j)$ denote a problem-specific residual and $\epsilon$ an inlier threshold. Under the data-generating parameter $\bfth^o$, a valid association is expected to satisfy $f_{\bfth^o}(\mathbf{x}_i, \mathbf{y}_j)<\epsilon$, whereas a spurious association does so with lower probability. In the 1-to-1 setting, the consensus maximization~(CM) mechanism~\cite{fischler1981random} scores a hypothesis by the number of consistent associations, or inliers. Under m-to-m association, however, multiple candidate edges can share endpoints, so a robust mechanism must account for the underlying graph structure.

	We represent candidate associations by a bipartite graph $\mathcal{G}=(V,E)$, with bipartition $V=S\cup T$, where $\bfx_i\in S$ and $\bfy_j\in T$. Each edge $(\bfx_i,\bfy_j)\in E$ denotes a candidate association. Given a hypothesis $\hat{\bfth}$, we identify inlier edges in $O(|E|)$ time and form the induced \textit{inlier subgraph} $\mathcal{G}_{\hat{\bfth}}=(V_{\text{in}},E_{\text{in}})$. A \textit{graph matching}
	$M(\mathcal{G}_{\hat{\bfth}})\subseteq E_{\text{in}}$ is a set of edges that share no vertices and hence obeys the 1-to-1 association constraint. Let $M^*_{\hat{\bfth}}$ denote a maximum-cardinality matching in $\mathcal{G}_{\hat{\bfth}}$, whose cardinality is the largest number of mutually compatible inlier edges under $\hat{\bfth}$. MCM scores $\hat{\bfth}$ by $|M^*_{\hat{\bfth}}|$ and estimates $\bfth^o$ by maximizing this score:
	\begingroup
	\setlength{\abovedisplayskip}{3pt}
	\setlength{\belowdisplayskip}{3pt}
	\setlength{\abovedisplayshortskip}{3pt}
	\setlength{\belowdisplayshortskip}{3pt}
	\begin{equation}\label{eqn::MCM formulation}
		\max_{\bfth}|M^*_{\bfth}|.
	\end{equation}
	\endgroup
	For $|E|$ candidate associations, evaluating a hypothesis $\hat{\bfth}$ with~\eqref{eqn::MCM formulation} requires $O(|E|)$ time to identify inliers and $\mathcal{O}(|E_{\text{in}}|\sqrt{|V_{\text{in}}|})$ time to find a maximum-cardinality matching via the Hopcroft--Karp algorithm~\cite{hopcroft1973n}.
	\subsection{A Likelihood Interpretation of MCM}
\begin{figure}[!htb]
		\centering
		\includegraphics[width=\columnwidth]{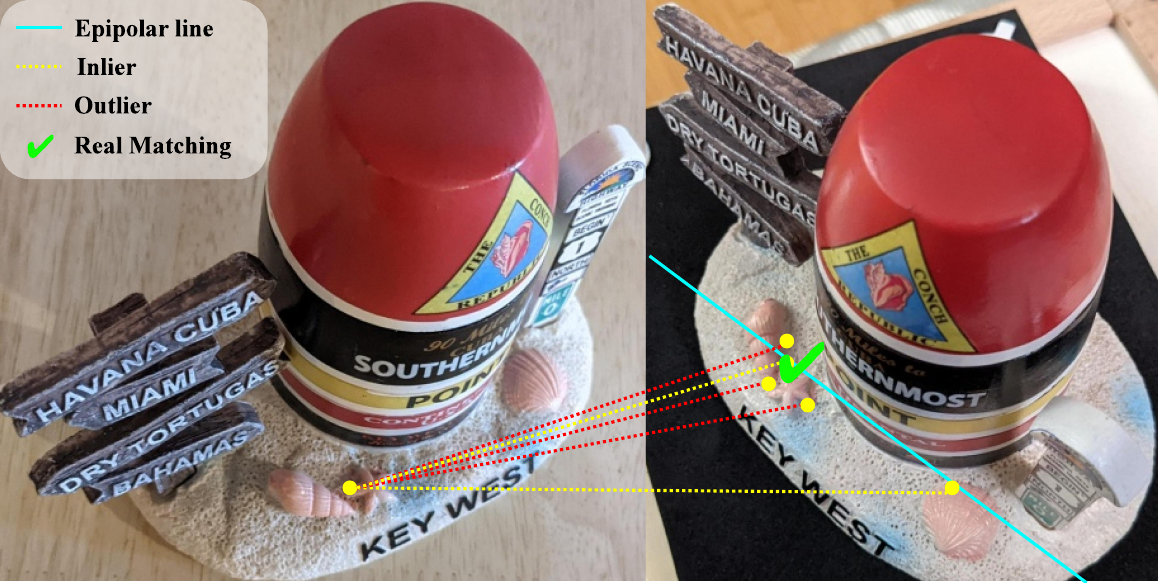}
		\caption{An inlier association under $\bfth^o$ is not necessarily real.}
		\label{fig::epipolar line}
	\end{figure}

	We introduce a working probabilistic model for m-to-m robust estimation. For each $\bfx_i\in S$, let $\mathcal{N}_{\bfx}(i)=\{j:(\bfx_i,\bfy_j)\in E\}$ denote the index set of its candidate neighbors in $T$. Likewise, we define $\mathcal{N}_{\bfy}(j)$ for each $\bfy_j\in T$. Our analysis builds on the following two assumptions.
	\begin{assumption}\label{assumption::real_association}
		For each datum $\bfx_i$ or $\bfy_j$:
		\begin{equation*}\label{eqn::real_association_priors}
			\begin{aligned}
			&\Pr[\exists j\in\mathcal N_{\bfx}(i)~\text{s.t.}~(\mathbf x_i,\mathbf y_j)
			\text{ is the real association}]=q_x^{(i)},\\
			&\Pr[\exists i\in\mathcal N_{\bfy}(j)~\text{s.t.}~(\mathbf x_i,\mathbf y_j)
			\text{ is the real association}]=q_y^{(j)},
			\end{aligned}
		\end{equation*}
	\end{assumption}
	\begin{assumption}\label{assumption::uniform_distribution}
		Under the ground truth parameter $\bfth^o$
		\begin{equation*}\label{eqn::uniform_residual_model}
			\begin{cases}
				&f_{\bfth^o}(\bfx_i,\bfy_j)\sim U(0,\epsilon)~~\text{if }(\bfx_i,\bfy_j)\text{ is a real association},\\
				&f_{\bfth^o}(\bfx_i,\bfy_j)\sim U(0,\frac{\epsilon}{\delta})~~\text{if }(\bfx_i,\bfy_j)\text{ is a false association},
			\end{cases}
		\end{equation*}
		where $\epsilon$ is the inlier threshold, and $\delta\in(0,1)$ characterizes how unlikely a spurious association is to appear as an inlier under the data-generating parameter.
	\end{assumption}
	In image matching, Assumption~\ref{assumption::real_association} reflects prior knowledge about keypoint-detection repeatability. Similarly, supervised methods assign probability to a ``dustbin'' when a keypoint is considered unmatched~\cite{sarlin2020superglue}. Assumption~\ref{assumption::uniform_distribution} extends the probabilistic justification~\cite{antonante2021outlier} of the 1-to-1 CM mechanism and is motivated by the following observation: a false association can appear to be an inlier under the real parameter, but with lower probability than a real one. For example, in relative pose estimation, a feature point far from the real association may be classified as an inlier simply because it lies near the epipolar line, as shown in Fig.~\ref{fig::epipolar line}.

	Based on the above two assumptions, we evaluate likelihood $l(\hat{\bfth})$ of a hypothesis $\hat{\bfth}$. In the association graph $\mathcal{G}=(V,E)$, each \textit{graph matching} $M_\tau(\mathcal{G}) \subseteq E$ corresponds to one possible configuration $\tau$ of real associations. Assuming a prior distribution $p_\tau$ over all possible configurations, the likelihood can be computed by marginalizing over $\tau$:
	\begingroup
	\setlength{\abovedisplayskip}{3pt}
	\setlength{\belowdisplayskip}{3pt}
	\setlength{\abovedisplayshortskip}{3pt}
	\setlength{\belowdisplayshortskip}{3pt}
	\begin{equation}\label{eqn::likelihood_calculation}
		l(\hat{\bfth})=\sum_{M_\tau(\mathcal{G})} p_\tau \cdot l(\hat{\bfth}|M_\tau(\mathcal{G})),
	\end{equation}
	\endgroup
	where the conditional likelihood $l(\hat{\bfth}|M_\tau(\mathcal{G}))$ is calculated under Assumption~\ref{assumption::uniform_distribution} and detailed in \textit{Supplementary}~\ref{append::likelihood_calculation}. We directly present the results here. If a matching $M_\tau(\mathcal{G})$ contains any outlier edge, $l(\hat{\bfth}|M_\tau(\mathcal{G})) = 0$; otherwise, the conditional likelihood writes $(\frac{\epsilon}{\delta})^{-|E|}(\frac{1}{\delta})^{|M_\tau(\mathcal{G})|}$. Thus, the likelihood~\eqref{eqn::likelihood_calculation} reduces to:
	\begingroup
	\setlength{\abovedisplayskip}{3pt}
	\setlength{\belowdisplayskip}{3pt}
	\setlength{\abovedisplayshortskip}{3pt}
	\setlength{\belowdisplayshortskip}{3pt}
	\begin{equation}\label{eqn::likelihood_2}
		l(\hat{\bfth})=\sum_{M_\tau(\mathcal{G}_{\hat{\bfth}})} p_{\tau}\cdot(\frac{\epsilon}{\delta})^{-|E|}(\frac{1}{\delta})^{|M_\tau(\mathcal{G}_{\hat{\bfth}})|},
	\end{equation}
	\endgroup
	where the sum is now restricted to graph matchings within the inlier graph $\mathcal{G}_{\hat{\bfth}}$. However, exactly computing~\eqref{eqn::likelihood_2} has two bottlenecks: (a) enumerating all graph matchings, and (b) properly adapting $p_\tau$ to incorporate prior knowledge (e.g., Assumption~\ref{assumption::real_association}). We instead resort to its dominant terms which admit a simple approximation.

	Let $N^*_{\hat{\bfth}}$ denote the number of maximum-cardinality
	matchings in $\mathcal{G}_{\hat{\bfth}}$. We assume that $\delta\ll1$ and
	$p_\tau$ is uniform, a matching of cardinality
	$|M_\tau(\mathcal{G}_{\hat{\bfth}})|<|M^*_{\hat{\bfth}}|$
	is suppressed relative to a maximum-cardinality matching by the
	factor
	$\delta^{|M^*_{\hat{\bfth}}|
	-|M_\tau(\mathcal{G}_{\hat{\bfth}})|}$.
	Retaining only the maximum-cardinality terms and taking logarithms
	therefore gives
	\begingroup
	\setlength{\abovedisplayskip}{3pt}
	\setlength{\belowdisplayskip}{3pt}
	\setlength{\abovedisplayshortskip}{3pt}
	\setlength{\belowdisplayshortskip}{3pt}
	\begin{equation}\label{eqn::log_likeli}
		\log l(\hat{\bfth})
		\approx
		\log N^*_{\hat{\bfth}}
		+|M_{\hat{\bfth}}^*|\log\!\left(\frac{1}{\delta}\right)
		+\mathrm{const.}
	\end{equation}
	\endgroup
	The derivation is provided in
	\textit{Supplementary}~\ref{append::likelihood_calculation}, and the additive
	constant in~\eqref{eqn::log_likeli} is independent of
	$\hat{\bfth}$. Further discarding the multiplicity term $N^*_{\hat{\bfth}}$ recovers the MCM score~\eqref{eqn::MCM formulation}.
	\subsection{Harmonic Consensus Maximization}
	Grounded in the same probabilistic perspective, we propose a real-valued robust score that requires only $O(|E|)$ time per evaluation, in contrast to $O(|E|+|E_{\rm in}|\sqrt{|V_{\rm in}|})$ for MCM. Our mechanism relies on the marginal probability $p_{i,j}$ assigned to each association $(\bfx_i,\bfy_j)$. Supervised matching models~\cite{sarlin2020superglue,lindenberger2023lightglue,wang2024efficient,jiang2024omniglue} usually output $p_{ij}$ through an optimal transport layer with differentiable Sinkhorn iterations~\cite{cuturi2013sinkhorn} or a dual-softmax operation. Since their effectiveness depends on co-training feature extraction and probability assignment, we cannot directly use them in our zero-shot setting. We detail our assignment algorithm in Section~\ref{section::marginal_prob_assign} and assume for now that $p_{i,j}$ is given.
	
	First, we unilaterally relax the 1-to-1 constraint and untangle different data points $\bfx_i$ from competing for a common match $\bfy_j$. Under this relaxation, the likelihood contributions $l_{\bfth}(\bfx_i)$ from different data points become independent, and the total likelihood simply equals $\prod_i l_{\bfth}(\bfx_i)$. Instead of marginalizing over configuration $\tau$ as in~\eqref{eqn::likelihood_calculation}, we define the following hidden variable $\iota_i$ for each $\bfx_i$:
	\begin{equation*}\label{eqn::assignment_variable}
	\begin{cases}
		&\iota_i = 0\text{ if the real association is not within } \{\bfy_j|j\in\mathcal{N}_{\bfx}(i)\},\\
		&\iota_i = j\text{ if }\bfy_j\text{ is the real association, for } j\in\mathcal{N}_{\bfx}(i).
	\end{cases}
	\end{equation*}
	The detailed derivation in \textit{Supplementary}~\ref{append::likelihood_calculation} gives:
	\begingroup
	\setlength{\abovedisplayskip}{3pt}
	\setlength{\belowdisplayskip}{3pt}
	\setlength{\abovedisplayshortskip}{3pt}
	\setlength{\belowdisplayshortskip}{3pt}
	\begin{equation}\label{eqn::one_sided_likelihood_main}
	l_{\hat{\bfth}}(\bfx_i)=(1-q_{x}^{(i)})(\frac{\delta}{\epsilon})^{|\mathcal{N}_{\bfx}(i)|}[1+C_x^{(i)}w_i(\hat{\bfth})],
	\end{equation}
	\endgroup
	where $C_x^{(i)}=\frac{q_{x}^{(i)}}{1-q_{x}^{(i)}}\frac{1}{\delta}$, and 
		$
		w_i(\hat{\bfth})=\frac{\sum_{j\in\mathcal{N}_{\bfx}(i)} \bfone[f_{i,j}<\epsilon]p_{i,j}}{q_x^{(i)}}
		$
	is the summed weight of $\bfx_i$'s inliers. Intuitively, a larger $C_x^{(i)}$ indicates a less ambiguous setting. Next, we apply the same ``matching-one'' relaxation to the complementary side of the data~($\bfy_j$), following the same derivation. This introduces the corresponding constant $C_y^{(j)}$ and weight $w_j(\hat{\bfth})$. After subtracting constants and summing results from both directions, we arrive at a concise, symmetric formulation, \textit{Harmonic Consensus Maximization}~(HCM):
		\begingroup
		\setlength{\abovedisplayskip}{3pt}
		\setlength{\belowdisplayskip}{3pt}
		\setlength{\abovedisplayshortskip}{3pt}
		\setlength{\belowdisplayshortskip}{3pt}
		\begin{equation}\label{eqn::HCM}
			\max_{\bfth}\sum_i\log(1+C_x^{(i)}w_i(\bfth))+\sum_j\log(1+C_y^{(j)}w_j(\bfth)).
		\end{equation}
		\endgroup
	Compared with MCM, HCM has linear-time complexity and a real-valued score, at some accuracy cost from relaxing the 1-to-1 regularity. Section~\ref{section::ransac_algorithms} combines both mechanisms in a two-stage LO-RANSAC.
	
	\subsection{Marginal Probability Assignment}
	\label{section::marginal_prob_assign}
	To apply~\eqref{eqn::HCM}, we assign the marginal probability $p_{i,j}$ for each association $(\bfx_i,\bfy_j)$. Heuristically, we introduce the hyperparameter $q$ as a reference value of $q_x^{(i)}$ and $q_y^{(j)}$ in Assumption~\ref{assumption::real_association} and assign $p_{i,j}$ under a principle of \textbf{uniformity}: retained associations are not differentiated by cosine similarity. Specifically, we assign $p_{i,j}$ as
	\begingroup
	\setlength{\abovedisplayskip}{3pt}
	\setlength{\belowdisplayskip}{3pt}
	\setlength{\abovedisplayshortskip}{3pt}
	\setlength{\belowdisplayshortskip}{3pt}
	\begin{equation}\label{eqn::uniform_assign}
		p_{i,j} = \frac{q}{2|\mathcal{N}_{\bfx}(i)|}+\frac{q}{2|\mathcal{N}_{\bfy}(j)|}.
	\end{equation} 
	\endgroup
	The resulting $q_x^{(i)}$ and $q_y^{(j)}$ thus approximate the reference $q$. In experiments, we set $q=0.3$ and $\delta=0.01$ for all evaluated datasets. As shown in \textit{Supplementary}~\ref{append::ablation_and_sensitivity}, the two-stage estimator's accuracy is relatively insensitive to both choices.
	\section{Two-Stage Many-to-Many LO-RANSAC}
	\label{section::ransac_algorithms}
\begin{figure}[!t]
		\centering
		\includegraphics[width=\linewidth]{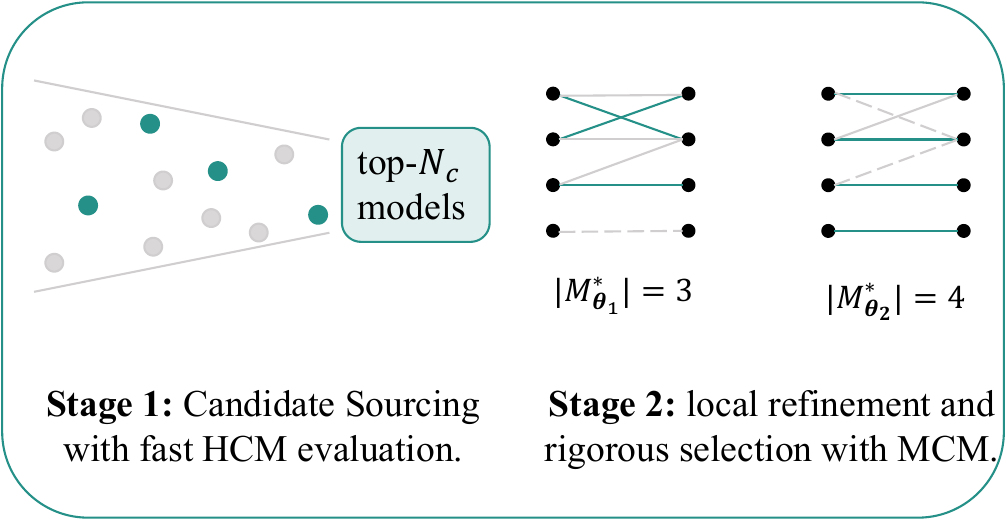}
		\caption{Two-stage HCM--MCM LO-RANSAC.}
		\label{fig::loransac}
	\end{figure}

	\begin{algorithm}[!t]
		\caption{Two-Stage M-to-M LO-RANSAC}\label{alg::m2m_RANSAC}
		\small
		\SetKwInOut{Input}{Input}\SetKwInOut{Output}{Output}
		\SetKwFunction{Solve}{Solve}\SetKwFunction{Sample}{Sample}\SetKwFunction{HCMScore}{HCMScore}\SetKwFunction{LocalRefine}{LocalRefine}\SetKwFunction{FinalPolish}{FinalPolish}\SetKwFunction{TryInsert}{TryInsert}
		\Input{association graph $\mathcal{G}$; residual $f_{\bfth}$ with standard and relaxed inlier thresholds $\epsilon$ and $\tilde{\epsilon}$; priors $\{p_{i,j}\}$ and constants $\{C_x^{(i)},C_y^{(j)}\}$; minimal solver \Solve{$\cdot$} of size $m$; bounds $T_{\min},T_{\max}$, $A_{\max}$; nominal stopping target $\eta$, safety factor $\kappa$; pool capacity $N_c$}
		\Output{model $\bfth^\ast$}
		\tcc{\small Stage 1: Candidate Sourcing}
		$s^\ast,\bar{s}\leftarrow-\infty$;\ \ $\bar\rho\leftarrow0$;\ \ $T^\ast\leftarrow T_{\max}$;\ \ $\mathcal{C}\leftarrow\emptyset$ \;
		\For{$t\leftarrow 1$ \KwTo $T^\ast$}{
			$U_c\leftarrow$ \Sample{$\mathcal{G}$,$m$,$A_{\max}$}\;
			\lIf{$U_c=\emptyset$}{\textbf{continue}}
			$\Theta\leftarrow$ \Solve{$U_c$}
			\lIf{$\Theta=\emptyset$}{\textbf{continue}}
			\ForEach{$\bfth\in\Theta$}{
				$(s_{\bfth},\rho_{\bfth})\leftarrow$ \HCMScore{$\bfth,\epsilon,p_{i,j},C_x^{(i)},C_y^{(j)}$}\;
			}
			\tcp{\small Maintain the top-$N_c$ list}
			$\hat{\bfth}\leftarrow\arg\max_{\bfth\in\Theta}s_{\bfth}$, \TryInsert{$\mathcal{C}$, $(\hat{\bfth},s_{\hat{\bfth}})$}\;
			\If(\tcp*[h]{\small LO gate}){$s_{\hat{\bfth}}>\bar{s}$ \textnormal{or} $\rho_{\hat{\bfth}}\ge\bar\rho$}
			{
				$\bar{s}\leftarrow\max(\bar{s},s_{\hat{\bfth}})$;\ \ $\bar\rho\leftarrow\max(\bar\rho,\rho_{\hat{\bfth}})$\;
				$\tilde{\bfth}\leftarrow$ \LocalRefine{$\hat{\bfth},\tilde{\epsilon}$}\;
				$(s_{\tilde{\bfth}},\rho_{\tilde{\bfth}})\leftarrow$ \HCMScore{$\tilde{\bfth},\epsilon,p_{i,j},C_x^{(i)},C_y^{(j)}$}\;
				\lIf{$s_{\tilde{\bfth}}>s_{\hat{\bfth}}$}{$(\hat{\bfth},s_{\hat{\bfth}},\rho_{\hat{\bfth}})\leftarrow(\tilde{\bfth},s_{\tilde{\bfth}},\rho_{\tilde{\bfth}})$}
			}
			\If{$s_{\hat{\bfth}}>s^\ast$}
			{
				$s^\ast\leftarrow s_{\hat{\bfth}}$;\ \ $I_{\hat{\bfth}}\leftarrow |E_{\rm in}(\hat{\bfth})|$;\ $\widehat p_{\hat{\bfth}}\leftarrow\binom{I_{\hat{\bfth}}}{m}/\binom{|E|}{m}$;\ \ $T^\ast\leftarrow\operatorname{clip}_{[T_{\min},T_{\max}]}\!\left\lceil\frac{\kappa\log(1{-}\eta)}{\log(1{-}\widehat p_{\hat{\bfth}})}\right\rceil$\;
			}
		}
		\tcc{\small Stage 2: Refine and Selection}
		\lIf{$\mathcal C=\emptyset$}{\Return $\mathrm{failure}$}
		\ForEach{$(\bfth,s_{\bfth})\in\mathcal{C}$}{
			$\bfth\leftarrow$ \LocalRefine{$\bfth,\tilde{\epsilon}$} \tcp*{\small local optimization}
			$c_{\bfth}\leftarrow|M^\ast_{\bfth}|$\tcp*{\small maximum matching, Eq.~\eqref{eqn::MCM formulation}}
		}
		$\bfth^\ast\leftarrow\arg\max_{\bfth}c_{\bfth}$, ties broken by the Stage~1 score $s_{\bfth}$\;
		\Return \FinalPolish{$\bfth^\ast,\epsilon$}\;
	\end{algorithm}
	We condense our theoretical findings in a two-stage LO-RANSAC~(Algorithm~\ref{alg::m2m_RANSAC}), which combines the complementary strengths of two robust mechanisms: HCM evaluates hypotheses in linear time and supplies a real-valued score but relaxes the 1-to-1 regularity, whereas MCM enforces this regularity but is more costly per evaluation. We therefore explore a large hypothesis space with HCM, distill it into a small top-$N_c$ shortlist, and re-rank this shortlist with MCM. Figure~\ref{fig::loransac} summarizes the details below.

	\textbf{Constrained minimal sampling.} Because each feature may belong to several associations, a naively sampled minimal set can yield a degenerate configuration. The \texttt{Sample} subroutine of Algorithm~\ref{alg::m2m_RANSAC} enforces the 1-to-1 constraint \emph{within} each sample~(pseudocode in \textit{Supplementary}~\ref{append::m2m_loransac}).

\textbf{Stage 1: exploration with HCM.} Each minimal sample is passed to~\texttt{Solve}$(\cdot)$, which may return multiple model roots. HCM evaluates every root in $O(|E|)$ time, and the highest-scoring root $\hat{\bfth}$ becomes the \emph{raw seed} for the current iteration. Before local optimization~(LO), $\hat{\bfth}$ is inserted into a bounded pool $\mathcal{C}$ that retains the $N_c$ highest-scoring raw seeds for Stage~2. LO is triggered when $\hat{\bfth}$ improves either the running-best HCM score or the running maximum edge-inlier ratio $\bar{\rho}$. If the refined seed attains the highest HCM score so far, it updates the adaptive iteration cap $T^\ast$. Let $I_{\hat{\bfth}}=|E_{\rm in}(\hat{\bfth})|$. We use $\widehat p_{\hat{\bfth}}=\binom{I_{\hat{\bfth}}}{m}/\binom{|E|}{m}$ as an analogue of the all-inlier-sample probability~\cite{chum2003locally} and set $T^\ast$ by clipping $\lceil\kappa\log(1-\eta)/\log(1-\widehat p_{\hat{\bfth}})\rceil$ to $[T_{\min},T_{\max}]$.

\textbf{Stage 2: selection with MCM.} Each candidate hypothesis is locally optimized once and then scored by~\eqref{eqn::MCM formulation}. Each LO step re-fits $\bfth$ over its inliers, gathered within the relaxed threshold $\tilde{\epsilon}$, under a robust loss. An association $(\bfx_i,\bfy_j)$ receives the weight $\omega_{i,j}=1/d_i+1/d_j$, where $d_i$ and $d_j$ are its endpoint degrees in the current inlier subgraph. We select the refined model with the largest cardinality, breaking ties by the real-valued HCM score retained from Stage~1. Before output, we polish the best refined model once more with \textsc{FinalPolish}, which uses the strict threshold $\epsilon$ and a robust kernel different from that of \textsc{LocalRefine}.

\section{Experiments}\label{sec::experiments}
\begin{figure*}[!t]
	\centering
	\includegraphics[width=0.98\textwidth]{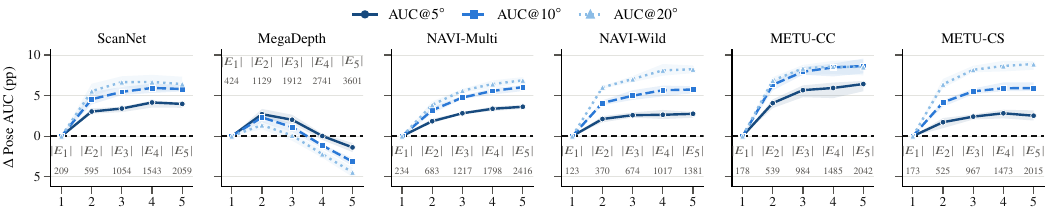}
	\caption{Fixed-proposal $K$-sweep over six datasets. MKNN produces nested graphs $E_1\subseteq\cdots\subseteq E_5$. Minimal samples are drawn from $E_1$ for all $10^5$ iterations, whereas $E_K$ is used for scoring, refinement, and re-ranking in Algorithm~\ref{alg::m2m_RANSAC}. Curves show the mean $\Delta\mathcal{A}$ relative to $K=1$ over five pair-specific random seeds; shaded bands show the min--max range. Each panel reports the pair-level median $|E_K|$.}
	\label{fig::controlled_validation}
\end{figure*}

The experiments test the preserve-then-resolve design along the evidence chain established above for relative-pose estimation. \textbf{Q1---Candidate preservation:} can non-rank-one candidates improve downstream pose estimation when RANSAC minimal proposals are fixed? \textbf{Q2---Ambiguity resolution:} on identical m-to-m graphs, are HCM and MCM effective, efficient, and complementary? \textbf{Q3---Shared ambiguity mechanism:} does the same fixed pipeline improve representation-matched rank-one baselines across the evaluated general-purpose representations? \textbf{Q4---Contextual comparison:} where does the resulting zero-shot DINOv3 pipeline stand relative to supervised matching systems across the tested image domains? Q1--Q3 test our central mechanism; Q4 asks whether this alignment yields meaningful system-level performance.
\subsection{Evaluation Settings}\label{sec::evaluation_settings}
\textbf{Datasets.} We evaluate relative pose estimation on six test sets. ScanNet-1500 and MegaDepth-1500 are standard indoor and outdoor RGB benchmarks~\cite{dai2017scannet,li2018megadepth}. NAVI-Multi uses fixed-scene multi-view capture, whereas NAVI-Wild introduces broader variation in backgrounds, illumination, and cameras; both cover the same object categories~\cite{jampani2023navi}. METU-CC and METU-CS probe visible--thermal matching~\cite{tuzcuouglu2024xoftr}. The 1,500-pair NAVI-Wild pose split is globally image-disjoint from the diagnostic split in Section~\ref{section::DINO_diagnosis}.

\textbf{Metric and common protocol.} The pose error is the larger of the rotation and translation-direction errors. We report normalized \textit{Pose AUC}@$5^\circ/10^\circ/20^\circ$~(\%) as an ordered triplet $a/b/c$ and denote changes in percentage points~(pp) by $\Delta\mathcal{A}$. All 1-to-1 and m-to-m LO-RANSAC variants use a 1-pixel threshold. SuperPoint-based methods share the same off-the-shelf detections and the per-image cap $N_{\rm kp}=2048$. Unless otherwise noted, stochastic comparisons use five pair-specific random seeds: for pair $i\in\{1,\ldots,N\}$ and repetition $b\in\{0,\ldots,4\}$, the effective seed is $s_{i,b}=i+b$.

\textbf{Question-specific configurations.} For each evaluated general-purpose representation, we bilinearly sample its patch-descriptor map at the same set of at most 2048 SuperPoint~\cite{detone2018superpoint} keypoints in each image. Let $N_{\rm a}$ and $N_{\rm e}$ denote the per-pair association and estimator-input caps, respectively. Q1 uses uncapped MKNN graphs and a fixed budget of $10^5$ iterations of Algorithm~\ref{alg::m2m_RANSAC}. Q2--Q4 use a common practical configuration for the ambiguity-aware pipeline, which incrementally expands the MKNN graph up to $K=5$, subject to the pre-GMS cap $N_{\rm a}=2048$, and then applies an m-to-m adaptation of Grid-Based Motion Statistics~(GMS)~\cite{bian2017gms} to remove spatially inconsistent candidates. Where applicable, the pose estimators in Q2--Q4 use $N_{\rm e}=1024$, $T_{\min}/T_{\max}=10^3/10^5$, and $\eta=0.9999$. We use pool capacity $N_c=100$ for Algorithm~\ref{alg::m2m_RANSAC} throughout Q1--Q4. Further details are provided in \textit{Supplementary}~\ref{append:exp_details}.


\subsection{Design-Level Evaluation}
\textbf{Q1---Candidate preservation.} \emph{Controlled protocol.} We form nested MKNN graphs $E_1\subseteq\cdots\subseteq E_5$ but draw minimal samples from $E_1$ for every $K$. Only probability assignment, HCM scoring, refinement, and MCM re-ranking use $E_K$, fixing the minimal-proposal distribution while varying the available evidence. For each of the five pair-specific random seeds, we run $10^5$ iterations per image pair without GMS or the $N_{\rm a}/N_{\rm e}$ caps. We then score the ground-truth pose $\bfth_{\rm gt}$ with the same $E_K$ and HCM parameters and record whether it would enter the top-100 raw-seed pool.

\emph{Pose estimation result.} Figure~\ref{fig::layer_selection} establishes the association-level premise, and Figure~\ref{fig::controlled_validation} tests its downstream gains. As $K$ increases from 1 to 5, dataset-averaged \textit{Pose AUC} rises from $11.64/22.68/36.31$ to $14.63/27.52/42.07$. Across the five pair-specific random seeds, the componentwise ranges of $\Delta\mathcal{A}$ w.r.t $K=1$ are $[2.69,3.23]/[4.53,5.01]/[5.41,5.95]$~pp. Retaining multiple candidates improves pose accuracy on five of six datasets. MegaDepth has the most cluttered association graphs and the accuracy declines from $K=3$ to $K=5$.

\emph{Reference-hypothesis result.} The dataset-averaged top-100 inclusion rate rises from $11.2\%$ at $K=1$ to $15.7\%$ at $K=5$, whereas MegaDepth follows $55.1/55.9/52.1/47.4/42.5\%$ over $K=1,\ldots,5$. Comparing the reference pose with the fixed raw Stage-1 pool measures its HCM separability before MCM re-ranking and final refinement. Together with the earlier correspondence diagnostic, these results expose a \textit{coverage versus clutter} trade-off: increasing $K$ preserves geometric evidence missed at rank one but also introduces spurious candidates. The two-stage estimator usually exploits the additional evidence, though the preferred size is dataset-dependent. \textit{Supplementary}~\ref{append::ablation_and_sensitivity} reports the per-dataset statistics.
\begin{figure}[!t]
	\centering
	\includegraphics[width=\columnwidth]{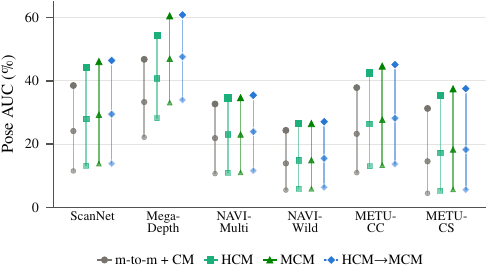}
	\caption{Robust-estimator comparison across six test sets, averaged over five pair-specific random seeds. All configurations receive the same graph for each image pair. Each vertical triplet connects the mean \textit{Pose AUC} values at $5^\circ/10^\circ/20^\circ$.}
	\label{fig::q2_estimator_comparison}
\end{figure}
\begin{table}[!tb]
	\centering
	\caption{Runtime on an AMD Ryzen~9 9950X. We report mean scorer time per evaluation and complete runtime per image pair.}
	\label{table::runtime}
	\small
	\setlength{\tabcolsep}{2.1pt}
	\begin{tabular}{l|cc|rrrr}
		\hline
		\multirow{2}{*}{Dataset} & \multicolumn{2}{c|}{$\mu$s\,/\,eval} & \multicolumn{4}{c}{runner ms\,/\,pair}\\
		& HCM & MCM & CM & HCM & MCM & H$\rightarrow$M\\
		\hline
		ScanNet-1500   & 4.7 & 6.3 & 787 & 1111 & 1365 & 1128\\
		MegaDepth-1500 & 6.5 & 9.4 & 387 & 557  & 771  & 592\\
		NAVI-Multi     & 4.5 & 5.6 & 760 & 1021 & 1183 & 1036\\
		NAVI-Wild      & 4.1 & 5.4 & 782 & 1006 & 1219 & 1018\\
		METU-CC        & 4.4 & 5.7 & 798 & 1137 & 1350 & 1158\\
		METU-CS        & 4.3 & 5.5 & 778 & 1103 & 1299 & 1117\\
		\hline
	\end{tabular}
\end{table}

\textbf{Q2---Ambiguity resolution.} On identical per-pair m-to-m graphs, we compare HCM-only, MCM-only, two-stage HCM$\rightarrow$MCM, and conventional CM. HCM-only and MCM-only use one scoring mechanism throughout, without candidate pooling or re-ranking. CM applies PoseLib's LO-RANSAC~\cite{PoseLib} to the same graph but treats its edges as independent putative correspondences. Figure~\ref{fig::q2_estimator_comparison} shows that HCM-only, MCM-only, and HCM$\rightarrow$MCM each outperform CM in all 18 dataset--threshold cells. The dataset-averaged $\Delta\mathcal{A}$ of the two-stage estimator over CM is $3.29/5.30/6.82$~pp, $1.41/2.07/2.46$~pp over HCM-only and $0.36/0.45/0.43$~pp over MCM-only.

Table~\ref{table::runtime} shows the computational trade-off. One MCM evaluation costs $1.25$--$1.46\times$ one HCM evaluation, and the complete two-stage runner is faster than MCM-only on every dataset and adds only $1$--$6\%$ runtime over HCM-only. MegaDepth combines the highest per-evaluation HCM/MCM costs with the lowest per-pair runner times: larger graphs make each evaluation more expensive, while a higher inlier ratio reduces adaptive RANSAC iterations. CM remains fastest but ignores the bipartite graph structure. Overall, HCM supports efficient exploration, while MCM enforces the 1-to-1 constraint during selection.

\textbf{Q3---Shared ambiguity across representations.} We test corrected layer-19 DINOv3 ViT-L/16~\cite{simeoni2025dinov3,cuttano2026insid3}, raw final-layer DINOv2 ViT-L/14-register~\cite{oquab2023dinov2}, final-layer V-JEPA~2.1 ViT-L/16~\cite{murlabadia2026vjepa2_1}, and final-layer SigLIP~2 So400m/16 NaFlex~\cite{tschannen2025siglip2} features. V-JEPA~2.1 and SigLIP~2 broaden the evaluation beyond DINO-style representations. We additionally include native SuperPoint descriptors as a task-specific control trained for local feature matching. Within each representation, the rank-one baseline uses MNN associations followed by PoseLib CM, whereas the preserve-then-resolve pipeline uses the common practical configuration defined in Section~\ref{sec::evaluation_settings}. Downstream association and estimation hyperparameters remain fixed across representations without model-specific tuning.

\begin{figure*}[!htbp]
	\centering
	\includegraphics[width=\textwidth]{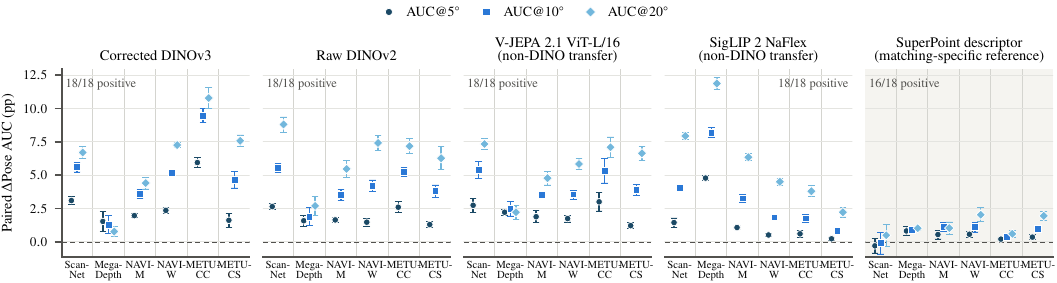}
	\caption{Within-representation improvement of preserve-then-resolve over MNN+PoseLib CM. Points and error bars show the mean and sample standard deviation of $\Delta\mathcal{A}$ over five pair-specific random seeds; panel text counts positive dataset--threshold means out of 18. The first four panels evaluate general-purpose representations, and SuperPoint provides the task-specific boundary control.}
	\label{fig::q3_representation_compatibility}
\end{figure*}

Figure~\ref{fig::q3_representation_compatibility} shows positive mean $\Delta\mathcal{A}$ in all 18 dataset--threshold cells for every general-purpose representation. Dataset-averaged gains are $2.75/4.96/6.25$~pp for DINOv3, $1.88/4.03/6.31$~pp for DINOv2, $2.14/4.03/5.65$~pp for V-JEPA~2.1, and $1.44/3.32/6.11$~pp for SigLIP~2. Under fixed downstream settings, this shared pattern supports a common, exploitable ambiguity mechanism: general-purpose features retain useful candidates beyond rank one, and geometric graph reasoning can resolve them. SuperPoint yields smaller and less uniform gains~($0.37/0.73/1.19$~pp; 16/18 positive cells), including negative means on ScanNet at $5^\circ$ and $10^\circ$. This control further shows that general-purpose features leave more rank ambiguity for multi-candidate reasoning. \textit{Supplementary}~\ref{append::ablation_and_sensitivity} gives full protocol and statistics.

\begin{table*}[!b]
	\centering
	\caption{Camera relative-pose accuracy~(\textit{Pose AUC}@$5^\circ\!/10^\circ\!/20^\circ$, \%). \textit{Ours} uses a practical DINOv3 preserve-then-resolve pipeline. \textbf{Bold} and underlining mark the best and second-best methods; modality-specialist XoFTR$^\dagger$ is excluded from the ranking.}
	\label{table::relative_pose_accuracy}
	\setlength{\tabcolsep}{3pt}
	\begin{tabular}{c|cccccc}
		\hline
		\multirow{2}{*}{Method} & ScanNet-1500&MegaDepth-1500 & NAVI-Multi & NAVI-Wild & METU-CC  & METU-CS \\
		& \multicolumn{6}{c}{\textit{Pose AUC}@ $5^\circ$ / $10^\circ$ / $20^\circ$}\\
		\hline
		SP    &13.5 / 27.3 / 42.1&43.5 / 57.6 / 68.3&11.6 / 22.1 / 31.2&2.4 / 6.5 / 12.0&0.0 / 0.2 / 0.2&0.0 / 0.1 / 0.3\\
		SP + LG  &\underline{19.2} / \underline{36.3} / \underline{53.1}&\underline{59.4} / \underline{73.5} / \underline{83.9}&\underline{15.4} / \underline{28.9} / \underline{39.2}&5.7 / 13.7 / 21.6 &4.2 / 8.6 / 13.8&\underline{5.8} / 11.6 / 18.3\\
		ELoFTR  &\textbf{21.5} / \textbf{39.1} / \textbf{54.9}&\textbf{66.1} / \textbf{78.8} / \textbf{87.6}&14.6 / 27.5 / 37.7& \textbf{7.3} / \textbf{15.6} / \underline{23.0} & \underline{8.6} / \underline{17.0} / \underline{28.7}&\textbf{7.9} / \underline{16.7} / \underline{29.1}\\
		OmniGlue  &16.2 / 32.3 / 48.2&56.2 / 70.9 / 81.5&14.6 / 27.4 / 37.4&6.1 / 13.1 / 20.6&2.9 / 6.2 / 13.1& 4.4 / 10.5 / 19.0\\
		MASt3R    &13.6 / 27.4 / 41.9&19.2 / 31.8 / 46.4&\textbf{18.7} / \textbf{40.3} / \textbf{58.1} & 4.0 / 9.6 / 15.9&0.7 / 3.0 / 8.6&0.8 / 3.3 / 9.4\\
		XoFTR$^\dagger$ &16.8 / 31.6 / 45.9&63.7 / 77.2 / 86.6&12.4 / 24.0 / 33.1&5.4 / 11.4 / 17.2&30.8 / 47.3 / 61.3&18.5 / 34.3 / 50.8\\
		\hline
		Ours&13.9 / 29.4 / 46.4 &34.0 / 47.5 / 60.8 &11.6 / 23.9 / 35.4 &\underline{6.4} / \underline{15.5} / \textbf{27.1} &\textbf{13.7} / \textbf{28.2} / \textbf{45.1} &5.6 / \textbf{18.3} / \textbf{37.5}\\
		\hline
	\end{tabular}
\end{table*}

\subsection{Contextual System-Level Evaluation}
\textbf{Q4---Comparison with supervised matchers.} \emph{Compared systems.} SuperPoint~\cite{detone2018superpoint} is a sparse detector--descriptor baseline; LightGlue~\cite{lindenberger2023lightglue} performs learned sparse matching over SuperPoint features; OmniGlue~\cite{jiang2024omniglue} combines DINOv2 and SuperPoint features for generalizable matching; ELoFTR~\cite{wang2024efficient} uses detector-free coarse-to-fine refinement; and MASt3R~\cite{leroy2024grounding} predicts dense correspondences from learned 3D representations. All SuperPoint-based systems share our off-the-shelf detections. XoFTR~\cite{tuzcuouglu2024xoftr}, trained specifically for visible--thermal matching, is reported separately as a modality-specialist reference.

	\emph{Result.} Table~\ref{table::relative_pose_accuracy} compares complete systems; our row averages five pair-specific random seeds. Learned matching systems achieve the highest accuracy on the standard RGB benchmarks, with ELoFTR leading ScanNet-1500 and MegaDepth-1500. NAVI-Multi and NAVI-Wild share 35 object categories but differ in background, illumination, camera, and scene composition. From NAVI-Multi to NAVI-Wild, MASt3R changes from a dataset-leading $18.7/40.3/58.1$ to $4.0/9.6/15.9$, whereas ours changes from $11.6/23.9/35.4$ to $6.4/15.5/27.1$. On NAVI-Wild, our zero-shot pipeline ranks first among non-specialist methods at $20^\circ$ and second at $5^\circ/10^\circ$. Among methods not specifically trained for visible--thermal matching, it also leads on METU-CC and at $10^\circ/20^\circ$ on METU-CS. The modality-specialist XoFTR remains substantially stronger. 

\emph{Perspective.} Supervision equips the compared methods with the capability of intra-image context aggregation, inter-image reasoning, coarse-to-fine localization, and 3D priors~\cite{sarlin2020superglue,lindenberger2023lightglue,wang2024efficient,leroy2024grounding}. Without task-specific adaptation, our zero-shot pipeline instead relies on the compatability between genearl-purpose features and the preserve-then-resolve designs. Its competitiveness under several shifted domains is thus notable, whereas the remaining gap underscores the value of supervised intelligence. 

	\section{Conclusions}
	\label{section::conclusion}
	General-purpose visual features are semantically transferable but geometrically ambiguous: correct correspondences often fall below rank one in cosine similarity yet survive in a small candidate set. A preserve-then-resolve design retains multiple candidates in an m-to-m graph and defers the physical 1-to-1 decision to robust estimation. For m-to-m robust estimation, we introduce a probabilistic perspective to interpret MCM as a dominant-cardinality surrogate for likelihood, and propose a linear-time, real-valued scorer, HCM. A two-stage LO-RANSAC integrates the merits of both.

	Controlled relative-pose estimation experiments validate end-task gains our designs. Fixed-proposal and same-graph comparisons characterize the coverage--clutter trade-off and the benefit of ambiguity-aware estimation. A fixed zero-shot pipeline improves the 1-to-1 baselines of four general-purpose visual features across 18/18 dataset--threshold cells. Its DINOv3 realization also exceeds non-specialist supervised matchers in several shifted-domain evaluations, while trailing the leading methods within their training domains. Overall, preserving plausible candidates allows geometric reasoning to convert transferable visual features into usable correspondence evidence.

	{
		\small
		\bibliographystyle{ieeenat_fullname}
		\bibliography{references}
	}

\ifdefined\WACVMainOnly
\else
	\maketitlesupplementary
	\appendix

	\section{Representation and Association Diagnostics}
	\label{append::layer_selection}
	\label{append::operator_comparison}
	This section expands the feature-association study in Section~\ref{section::DINO_diagnosis}. We first detail the diagnostic split and geometric metric, then analyze the representation and association operator choices used in the main paper.

	\subsection{Diagnostic Protocol}
	\textbf{Image-pair sampling.} The released NAVI-Wild pair list~\cite{jampani2023navi} contains around 36,000 image pairs spanning 2,180 unique images. For each of the 35 objects, we partition its images into two pools; across objects, each pool contains around 1,090 images. We then form the diagnostic and pose-estimation pairs only within their respective pools. Consequently, the 1,500-pair diagnostic split used here and the separate 1,500-pair pose-estimation split used in Section~\ref{sec::experiments} are globally image-disjoint across all viewpoint-separation bins, rather than being merely pair-disjoint.

	Both the diagnostic and estimation splits contain 500 pairs in each viewpoint-separation bin $[0^\circ,40^\circ)$, $[40^\circ,80^\circ)$, and $[80^\circ,120^\circ)$. Within each bin, 25 objects contribute 14 pairs and the remaining 10 contribute 15; the extra-pair allocation rotates across bins. Corresponding object--bin cells in the two splits have identical $5^\circ$ angle histograms. The image partition additionally balances the NAVI train/validation tag, image resolution, and camera model. A deterministic pair-selection objective further balances same- versus cross-camera pairs and unordered camera-pair exposure. To improve view coverage, an image is reused at most four times within one bin and six times across all bins. In each split and viewpoint-separation bin, we swap the source and target images for exactly 250 of the 500 selected pairs, preventing systematic bias from image ordering. 
	
	\textbf{Feature extraction and association.} We resize the longer image edge to 1024 pixels and retain every complete, non-padding token on the native ViT/16 or ViT/14 grid. Descriptors are $\ell_2$-normalized, optionally debiased, and normalized again before neighbor search. Following~\cite{cuttano2026insid3}, debiasing projects DINOv3 features onto the null space of a rank-$s$ positional subspace estimated from a noisy, semantically uninformative image. We evaluate $s\in\{0,100,\ldots,600\}$, where $s=0$ denotes raw DINOv3 features. All diagnostics in this section are cap-free: top-$K$ search covers every complete token in the other image, including background, and the retained association graph is not truncated by an association-count cap.

	\subsection{Geometric Metric}
	We unproject patch centers into a common 3D frame using accurate depth and calibrated cameras provided by NAVI-Wild. A patch is \emph{valid} when its center lies within the object mask and has valid depth. At threshold $\tau$, a valid patch is \emph{eligible} if the other view contains at least one valid patch whose 3D distance is strictly below $\tau$ cm. A retained edge is correct only if both endpoints are valid and their 3D distance is strictly below $\tau$ cm. For each image in a pair, correspondence recall is the fraction of eligible patches incident to at least one correct retained association; the mean over the two images gives the recall for that pair. We then average recall over pairs within each object and weight the 35 object-level means equally within each viewpoint-separation bin. The reported recall is the arithmetic mean over $\tau\in\{1,2,5\}$ cm. Object masks, depth, and camera poses are used only to evaluate correspondence recall and never restrict descriptor search.

	\subsection{Representation Diagnostics}
	Figure~\ref{fig::rank_sweep_all_bins} complements the fixed-debiasing-rank comparison in Fig.~\ref{fig::layer_selection} by jointly varying the DINOv3 layer, debiasing rank $s$, and candidate count $K$. Each curve plots correspondence recall against the retained MKNN association count; points closer to the upper left provide a better recall versus association-count trade-off. All markers are direct observations for MKNN with $K=1,\ldots,8$. As viewpoint separation increases, the uncorrected $s=0$ curves fall increasingly below the corrected curves.

	To compare adjacent debiasing ranks independently of their differing association counts, we linearly interpolate each pair of curves at 201 uniformly spaced association budgets over their shared count range and compute the mean absolute recall gap. We then average this gap equally across layers and viewpoint-separation bins. Relative to the raw representation~($s=0$), the first corrected setting~($s=100$) improves recall by 6.52 percentage points~(pp) on average. Beyond this initial correction, the results are insensitive to rank: the interpolated recall curves for adjacent settings between $s=100$ and $s=600$ differ by only 0.18--0.42~pp on average, and no setting consistently dominates across layers or viewpoint-separation bins. We therefore use $s=200$ as a representative setting from this stable range, without claiming that it is uniquely optimal.

	\clearpage
	\twocolumn[{
		\begin{minipage}{\textwidth}
			\centering
			\includegraphics[width=0.95\textwidth]{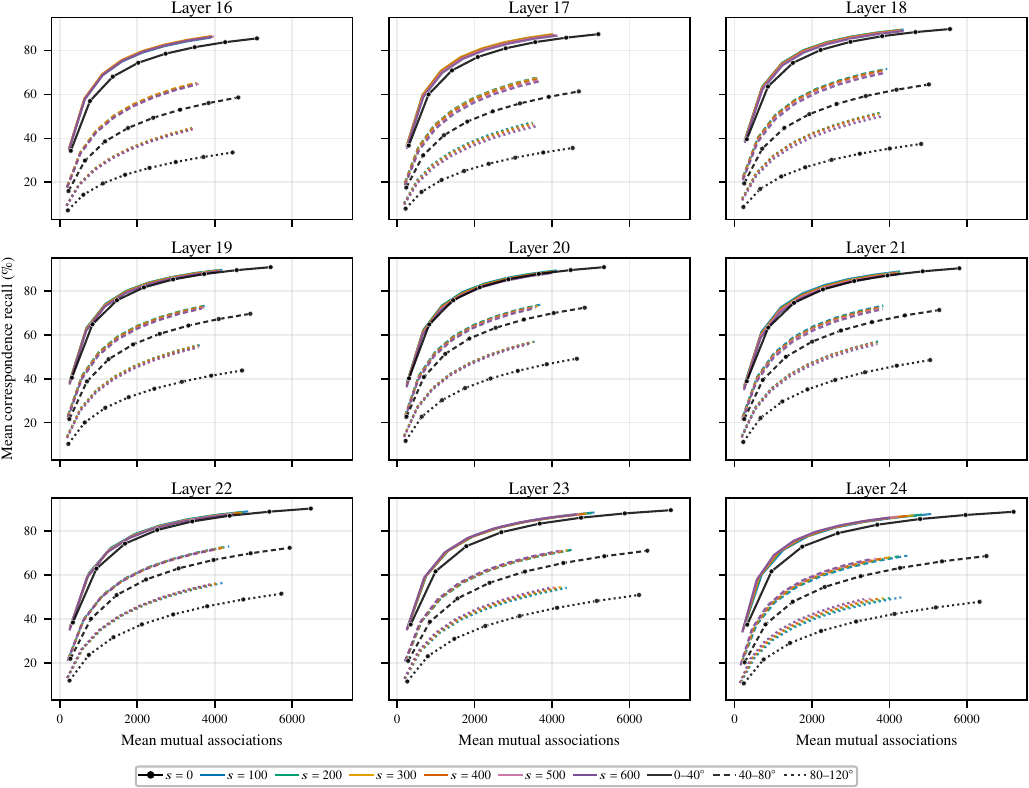}
			\captionof{figure}{Cap-free layer and debiasing-rank sweep on NAVI-Wild. Panels show DINOv3 ViT-L/16 layers; line styles indicate viewpoint-separation bins, and colors indicate $s=0,100,\ldots,600$. Curves connect MKNN observations for $K=1,\ldots,8$. Recall is the equally weighted mean over the 1, 2, and 5~cm thresholds; association counts are uncapped object-macro means. Corrected-rank curves cluster tightly, whereas the raw $s=0$ curves degrade most under large viewpoint changes.}
			\label{fig::rank_sweep_all_bins}
			\vspace{0.6em}
			\includegraphics[width=0.95\textwidth]{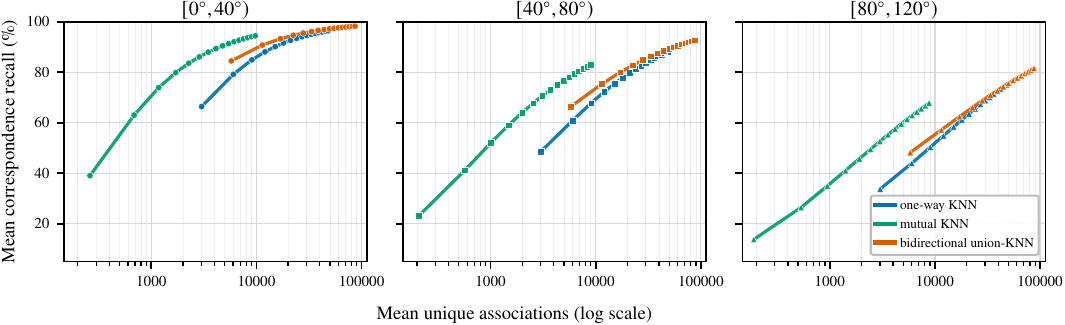}
			\captionof{figure}{Cap-free association operator comparison on NAVI-Wild. Panels show the three viewpoint-separation bins; curves connect observations for $K=1,\ldots,16$. Recall is the equally weighted mean over the 1, 2, and 5~cm thresholds; object-macro association counts use a logarithmic axis. MKNN consistently recovers more geometrically valid endpoints per retained association than one-way KNN or bidirectional union-KNN.}
			\label{fig::MKNN association}
		\end{minipage}
	}]
	\clearpage
	\twocolumn[{
		\begin{minipage}{\textwidth}
			\centering
			\includegraphics[width=\textwidth]{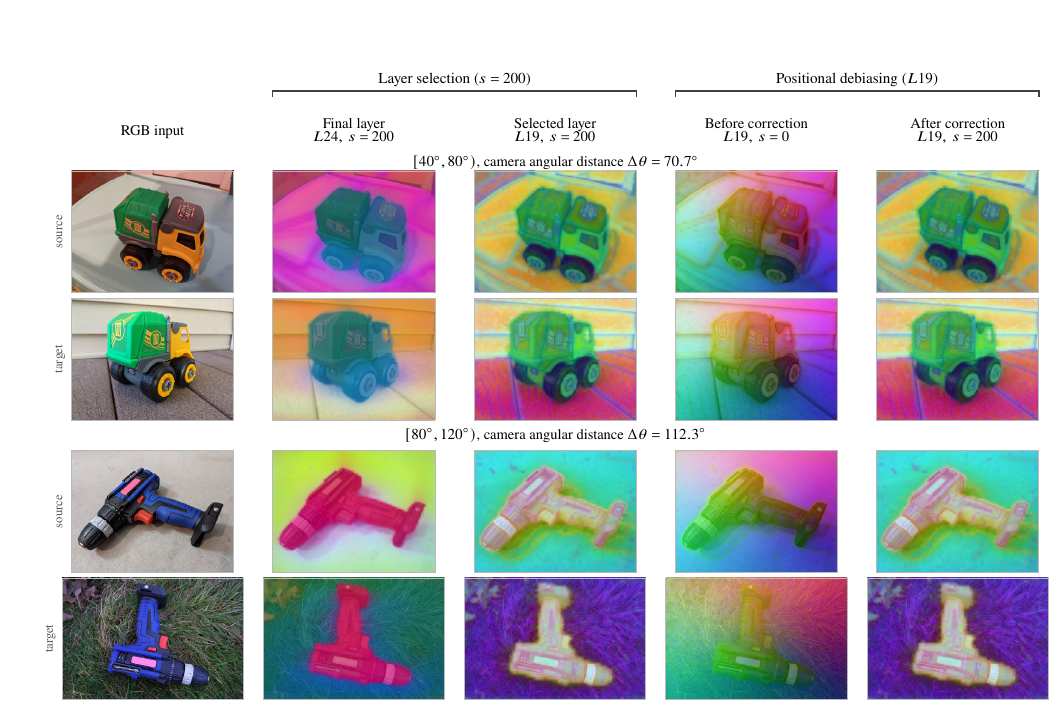}
			\captionof{figure}{Qualitative representation diagnostics on two image pairs from the NAVI-Wild diagnostic split. The comparisons isolate layer selection~(left bracket: $L24$ versus $L19$ at $s=200$) and positional debiasing~(right bracket: $s=0$ versus $s=200$ at $L19$). For each example and feature column, one PCA basis is fitted jointly to the source and target DINOv3 features; its first three components are robustly scaled to RGB and overlaid on the images. Colors are therefore comparable between the two views within a column, but not across columns. Debiasing removes the dominant coordinate-aligned color variation and reveals coherent object regions, while the selected intermediate layer better preserves cross-view object and part structure in these examples.}
			\label{fig::representation_qualitative}
		\end{minipage}
	}]

	To compare layers, we define a fixed-debiasing-rank empirical upper-left frontier separately for each viewpoint-separation bin. At an association budget $n$, this frontier is the highest recall attained by any tested layer/$K$ point using at most $n$ associations. The regret of an observed layer/$K$ point is the frontier recall at that point's association count minus its own recall, measured in pp; \emph{mean regret} averages these gaps equally over $K=1,\ldots,8$ and the three viewpoint-separation bins. At $s=200$, layer~19 has 0.42~pp mean regret, compared with 3.50~pp for the final layer~24. The same frontier-regret criterion selects the final layer~24 for raw DINOv2 features. These results support the representation choices in the main paper.
	
	Figure~\ref{fig::representation_qualitative} visualizes the two representation-side changes while controlling the other variable. We use one fixed diagnostic pair from each of the two larger-separation viewpoint bins, for which Fig.~\ref{fig::layer_selection} shows the clearest layer separation. Each title reports the corresponding viewpoint bin and camera angular distance. With the debiasing rank fixed at $s=200$, layer~19 produces more view-consistent object and part structure than layer~24. With the layer fixed at 19, the raw PCA maps are dominated by smooth spatial color variation tied to image coordinates, whereas debiasing reveals coherent object regions across the two views.

	Even after debiasing, correspondence recall at a fixed $K$ decreases as viewpoint separation increases. Retaining more candidates recovers a substantial fraction of the missed correspondences: for the selected layer-19 DINOv3 representation at $s=200$, increasing $K$ from 1 to 5 raises recall by 44.5~pp in $[0^\circ,40^\circ)$, 41.0~pp in $[40^\circ,80^\circ)$, and 32.0~pp in $[80^\circ,120^\circ)$. Raw DINOv2 shows the same trends. Thus, representation-side improvements do not remove the underlying geometric ambiguity: a geometrically valid counterpart may fall below rank one but remain in a small candidate set. This diagnostic establishes candidate availability and motivates the ambiguity-aware estimator in the main paper. Next, we compare m-to-m association operators.

	\subsection{Association Operator Diagnostics}
	We fix the DINOv3 representation to the selected layer~19 and debiasing rank $s=200$, and compare three cap-free operators at every $K=1,\ldots,16$~(Fig.~\ref{fig::MKNN association}). For \textit{one-way KNN}, we independently construct the two directional graphs and average their association counts and endpoint recalls. \textit{Bidirectional union-KNN} retains the deduplicated union of the two directional top-$K$ edge sets, whereas the adopted \textit{MKNN} retains their reciprocal intersection. Because equal $K$ produces very different graph sizes, we compare operating points by association count.

	At comparable or smaller association counts, MKNN with $K=6/6/7$ retains $2{,}863/2{,}546/3{,}000$ associations on average, compared with $3{,}013/3{,}015/3{,}016$ for one-way $K=1$, and improves recall by $19.72/19.15/18.89$~pp across the three viewpoint-separation bins. MKNN with $K=16$ exceeds one-way $K=8$ by $0.99/1.72/2.30$~pp while using only $40.5/37.4/36.5\%$ as many edges. Bidirectional union-KNN attains higher raw recall only after expanding to tens of thousands of edges. Within this cap-free dense-patch diagnostic, MKNN provides the strongest tested recall versus association-count trade-off.
	
	\section{Likelihood Derivations for MCM and HCM}
	\label{append::likelihood_calculation}
	This appendix expands on the likelihood derivations in Section~\ref{section::theory}: the dominant-cardinality approximation behind MCM and the one-sided relaxation behind HCM. 
	\subsection{MCM: Dominant-Cardinality Approximation}
	\label{append::dominant_matching}
	\begin{figure}[!htb]
		\centering
		\includegraphics[width=\columnwidth]{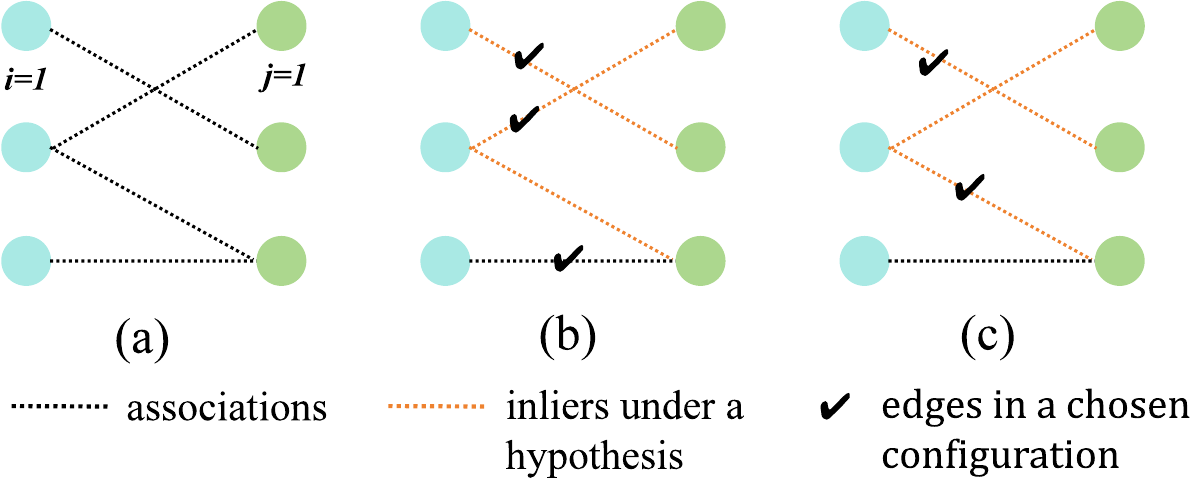}
		\caption{Toy configurations under an evaluated hypothesis $\hat{\bfth}$. (a)~Candidate association graph. Orange edges in (b)--(c) are inliers under $\hat{\bfth}$, and check marks indicate the selected 1-to-1 association subsets. Including an outlier as in (b) gives zero conditional likelihood, whereas a subset with all inliers as in (c) contributes a positive likelihood increasing with subset cardinality.}
		\label{fig::toy example}
	\end{figure}
	As defined in Section~\ref{section::theory}, a configuration $\tau$ selects a 1-to-1 subset of candidate associations, modeling its selected edges as real and the remaining edges as false. Under Assumption~\ref{assumption::uniform_distribution}, if the subset includes an outlier under $\hat{\bfth}$, its conditional likelihood is zero, as in Fig.~\ref{fig::toy example}(b). Otherwise, each selected edge contributes $1/\epsilon$ and each remaining edge contributes $\delta/\epsilon$; an all-inlier subset of cardinality $m$ therefore contributes $(1/\epsilon)^m(\delta/\epsilon)^{|E|-m}=(\epsilon/\delta)^{-|E|}(1/\delta)^m$. Because $0<\delta<1$, this contribution increases with $m$. 

	Let $m_\tau=|M_\tau(\mathcal{G}_{\hat{\bfth}})|$ be the cardinality of a contributing matching~(a 1-to-1 subset) and $m^*_{\hat{\bfth}}=|M^*_{\hat{\bfth}}|$ the maximum matching cardinality in the same inlier graph. The term associated with $\tau$ in Eq.~\eqref{eqn::likelihood_2} factors relative to a maximum-cardinality term as
	\begingroup
	\setlength{\abovedisplayskip}{3pt}
	\setlength{\belowdisplayskip}{3pt}
	\setlength{\abovedisplayshortskip}{3pt}
	\setlength{\belowdisplayshortskip}{3pt}
	\begin{equation}
	\label{eqn::dominant_term_factorization}
	p_\tau
	 \left(\frac{\epsilon}{\delta}\right)^{-|E|}
	 \left(\frac{1}{\delta}\right)^{m_\tau}=
	\left[
	p_\tau
	\left(\frac{\epsilon}{\delta}\right)^{-|E|}
	\left(\frac{1}{\delta}\right)^{m^*_{\hat{\bfth}}}
	\right]
	\delta^{m^*_{\hat{\bfth}}-m_\tau}.
	\end{equation}
	\endgroup
	A matching that is $r=m^*_{\hat{\bfth}}-m_\tau$ edges smaller is therefore suppressed by $\delta^r$. Assuming a uniform configuration prior $p_\tau=p$, the dominant contribution as $\delta\to0$ is the sum over the $N^*_{\hat{\bfth}}$ maximum-cardinality matchings:
	\begingroup
	\setlength{\abovedisplayskip}{3pt}
	\setlength{\belowdisplayskip}{3pt}
	\setlength{\abovedisplayshortskip}{3pt}
	\setlength{\belowdisplayshortskip}{3pt}
	\begin{equation}
		l(\hat{\bfth})
		\approx
		N^*_{\hat{\bfth}}\,p
		\left(\frac{\epsilon}{\delta}\right)^{-|E|}
	\left(\frac{1}{\delta}\right)^{m^*_{\hat{\bfth}}}.
	\label{eqn::dominant_matching_approximation}
	\end{equation}
	\endgroup
	Taking logarithms recovers Eq.~\eqref{eqn::log_likeli}. MCM retains the cardinality term $m^*_{\hat{\bfth}}\log(1/\delta)$ and drops the multiplicity term $\log N^*_{\hat{\bfth}}$. It therefore captures the dominant exponent but does not distinguish hypotheses with the same $m^*_{\hat{\bfth}}$.
	   
	\subsection{HCM: One-Sided Relaxation}
	HCM avoids enumerating graph matchings through a one-sided relaxation. For each $\bfx_i$, let $\mathcal{N}_{\bfx}(i)$ be its candidate indices and let $\iota_i\in\{0\}\cup\mathcal{N}_{\bfx}(i)$ denote either the no-match state~($\iota_i=0$) or the index $j$ of the real candidate $\bfy_j$~($\iota_i=j$). Define $p_{i,j}=\Pr(\iota_i=j)$ and $q_x^{(i)}=\sum_{j\in\mathcal{N}_{\bfx}(i)}p_{i,j}=\Pr(\iota_i\neq0)$, and abbreviate $f_{\hat{\bfth}}(\bfx_i,\bfy_j)$ as $f_{i,j}$. In the state $\iota_i=0$, all incident edges contribute the false-edge density; in the state $\iota_i=j$, edge~$j$ contributes the real-edge density when $f_{i,j}<\epsilon$, while the remaining edges contribute the false-edge density. Marginalizing over $\iota_i$ gives the contribution
	of $\bfx_i$ as
	\begingroup
	\setlength{\abovedisplayskip}{3pt}
	\setlength{\belowdisplayskip}{3pt}
	\setlength{\abovedisplayshortskip}{3pt}
	\setlength{\belowdisplayshortskip}{3pt}
	\begin{equation}\label{eqn::one_sided_likelihood_appendix}
		\begin{aligned}
		l_{\hat{\bfth}}(\bfx_i)
		&=(1-q_x^{(i)})\left(\frac{\delta}{\epsilon}\right)^{|\mathcal{N}_{\bfx}(i)|}\\
		&\quad+\sum_{j\in\mathcal{N}_{\bfx}(i)}p_{i,j}\bfone[f_{i,j}<\epsilon]
		\left(\frac{\delta}{\epsilon}\right)^{|\mathcal{N}_{\bfx}(i)|-1}\frac{1}{\epsilon}\\
		&=(1-q_{x}^{(i)})(\frac{\delta}{\epsilon})^{|\mathcal{N}_{\bfx}(i)|}[1+C_x^{(i)}w_i(\hat{\bfth})],
		\end{aligned}
	\end{equation}
	\endgroup
	where
	$C_x^{(i)}=\frac{q_x^{(i)}}{1-q_x^{(i)}}\frac{1}{\delta}$ and
	$w_i(\hat{\bfth})=\frac{1}{q_x^{(i)}}\sum_{j\in\mathcal{N}_{\bfx}(i)}\bfone[f_{i,j}<\epsilon]p_{i,j}$.
	Here, $w_i\in[0,1]$ is the fraction of the endpoint's candidate probability mass assigned to inliers, while $C_x^{(i)}$ scales the prior odds that a real match is present by the inverse false-inlier rate. After dropping the prefactor independent of $\hat{\bfth}$, multiplying over the relaxed $\bfx_i$ vertices and taking logarithms yields $\sum_i\log(1+C_x^{(i)}w_i(\hat{\bfth}))$. Repeating the construction from the $\bfy_j$ side and adding the two log scores gives the symmetric HCM objective in~\eqref{eqn::HCM}. Because the two directions reuse the same edges, this is a symmetric composite score rather than a joint log-likelihood.

	\section{M-to-M LO-RANSAC Algorithm}
	\label{append::m2m_loransac}
	Algorithm~\ref{alg::m2m_loransac} is the single-mechanism variant of the two-stage estimator, and it omits the candidate pooling and MCM re-ranking. Algorithm~\ref{alg::constrained_sample} gives the constrained \texttt{Sample} subroutine invoked by both this estimator and the two-stage estimator in Algorithm~\ref{alg::m2m_RANSAC}; it draws a 1-to-1 minimal subset from an m-to-m graph. Let $\rho_{\bfth}=|E_{\rm in}(\bfth)|/|E|$ denote the edge-inlier ratio. The HCM gate triggers when either the real-valued score improves or $\rho_{\bfth}$ matches or exceeds its running maximum; the MCM gate applies the analogous test to maximum-matching cardinality. The \textit{HCM} and \textit{MCM} baselines in Section~\ref{sec::experiments} instantiate these two choices.

	\begin{algorithm}[!htb]
		\small
		\caption{Constrained \texttt{Sample}}\label{alg::constrained_sample}
		\SetKwInOut{Input}{Input}\SetKwInOut{Output}{Output}
		\Input{association graph $\mathcal{G}$; sample size $m$; \\ attempt limit $A_{\max}$}
		\Output{1-to-1 association subset $U_c$ of size $m$, or $\emptyset$}
		$(S,T)\leftarrow$ active left/right vertex sets of $\mathcal{G}$\;
		\lIf{$|S|<m$ \textnormal{or} $|T|<m$}{\Return $\emptyset$}
		\For{$a\leftarrow 1$ \KwTo $A_{\max}$}{
			$U_c,V_c\leftarrow\emptyset$\;
			\ForEach{$\bfx_i\in\mathrm{RandPerm}(S)$}{
				\lIf{$|U_c|=m$}{\textbf{break}}
				choose a random cyclic start in $\mathcal N_{\bfx}(i)$ and take the first $\bfy_j\notin V_c$\;
				\lIf{such $\bfy_j$ exists}{$U_c\leftarrow U_c\cup\{(\bfx_i,\bfy_j)\}$;\ $V_c\leftarrow V_c\cup\{\bfy_j\}$}
			}
			\lIf{$|U_c|=m$}{\Return $U_c$}
		}
		\Return $\emptyset$\;
	\end{algorithm}

	\begin{algorithm}[!t]
		\small
		\caption{M-to-M LO-RANSAC~(Scorer $\mathcal{S}$)}\label{alg::m2m_loransac}
		\SetKwInOut{Input}{Input}\SetKwInOut{Output}{Output}
		\SetKwFunction{Solve}{Solve}\SetKwFunction{Sample}{Sample}\SetKwFunction{Score}{Score}\SetKwFunction{LocalRefine}{LocalRefine}\SetKwFunction{FinalPolish}{FinalPolish}\SetKwFunction{Gate}{ModeSpecificGate}
		\Input{association graph $\mathcal{G}$; residual $f_{\bfth}$ with inlier threshold $\epsilon$ and relaxed threshold $\tilde\epsilon$; scoring mechanism $\mathcal{S}\in\{HCM,MCM\}$ with priors $\{p_{i,j}\}$, constants $\{C_x^{(i)},C_y^{(j)}\}$; minimal solver \Solve{$\cdot$} of size $m$; bounds $T_{\min},T_{\max}$, $A_{\max}$; nominal stopping target $\eta$, safety factor $\kappa$}
		\Output{model $\bfth^\ast$}
		$s^\ast,\bar{s}\leftarrow-\infty$;\ \ $\bar\rho\leftarrow0$;\ \ $T^\ast\leftarrow T_{\max}$;\ \ $\bfth^\ast\leftarrow\mathrm{failure}$\;
		\For{$t\leftarrow 0$ \KwTo $T_{\max}-1$}{
			\lIf{$t>T_{\min}$ and $t>T^\ast$}{\textbf{break}}
			$U_c\leftarrow$ \Sample{$\mathcal{G}$, $m$, $A_{\max}$}\;
			\lIf{$U_c=\emptyset$}{\textbf{continue}}
			$\Theta\leftarrow$ \Solve{$U_c$}\tcp*{\small multiple roots}
			\lIf{$\Theta=\emptyset$}{\textbf{continue}}
			\lForEach{$\bfth\in\Theta$}{$(s_{\bfth},\rho_{\bfth})\leftarrow$ \Score{$\bfth$;\,$\mathcal{S}$}}
			$\hat{\bfth}\leftarrow\arg\max_{\bfth\in\Theta}s_{\bfth}$\;
			\If(\tcp*[h]{\small mechanism-specific LO gate}){\Gate{$\mathcal S,s_{\hat{\bfth}},\rho_{\hat{\bfth}};\bar{s},\bar\rho$}}
			{
				$\bar{s}\leftarrow\max(\bar{s},s_{\hat{\bfth}})$;\ \ $\bar\rho\leftarrow\max(\bar\rho,\rho_{\hat{\bfth}})$\;
				$\tilde{\bfth}\leftarrow$ \LocalRefine{$\hat{\bfth},\tilde\epsilon$}\; 
				$(s_{\tilde{\bfth}},\rho_{\tilde{\bfth}})\leftarrow$ \Score{$\tilde{\bfth}$;\,$\mathcal{S}$}\;
				\lIf{$s_{\tilde{\bfth}}>s_{\hat{\bfth}}$}{$(\hat{\bfth},s_{\hat{\bfth}},\rho_{\hat{\bfth}})\leftarrow(\tilde{\bfth},s_{\tilde{\bfth}},\rho_{\tilde{\bfth}})$}
			}
			\If{$s_{\hat{\bfth}}>s^\ast$}
			{
				$(\bfth^\ast,s^\ast)\leftarrow(\hat{\bfth},s_{\hat{\bfth}})$;\ \ $I_{\hat{\bfth}}\leftarrow\rho_{\hat{\bfth}}|E|$\;
				$\widehat p_{\hat{\bfth}}\leftarrow\binom{I_{\hat{\bfth}}}{m}/\binom{|E|}{m}$;\ \ $T^\ast\leftarrow\operatorname{clip}_{[T_{\min},T_{\max}]}\!\left\lceil\frac{\kappa\log(1{-}\eta)}{\log(1{-}\widehat p_{\hat{\bfth}})}\right\rceil$\;
			}
		}
		\Return \FinalPolish{$\bfth^\ast,\epsilon$}\;
	\end{algorithm}

		\section{\texorpdfstring{Experimental Protocols and\\Implementation Details}{Experimental Protocols and Implementation Details}}
		\label{append:exp_details}
		\textbf{Datasets.}
		\begin{itemize}
			\item \textbf{MegaDepth-1500} uses the LoFTR benchmark pairs~\cite{sun2021loftr} from two outdoor MegaDepth scenes~\cite{li2018megadepth}. We retain the conventional name, although its distributed ground-truth list contains 1,491 evaluable pairs.
			\item \textbf{ScanNet-1500} follows the SuperGlue benchmark~\cite{sarlin2020superglue} and contains 1,500 indoor pairs from ScanNet~\cite{dai2017scannet}.
			\item \textbf{NAVI-Multi} and \textbf{NAVI-Wild} each contain 1,500 object-centric pairs over the same 35 categories and three camera-angle bins~\cite{jampani2023navi}. Multi uses fixed-scene capture; Wild varies backgrounds, illumination, and cameras. The Wild pose split is image-disjoint from the diagnostic split constructed in Appendix~\ref{append::layer_selection}.
			\item \textbf{METU-CC} and \textbf{METU-CS} are from~\cite{tuzcuouglu2024xoftr} and contain 1,382 cloudy--cloudy and 1,208 cloudy--sunny visible--thermal pairs from six outdoor scenes.
		\end{itemize}

		\textbf{Evaluation metric.} We define relative-pose estimation error as the larger of the sign-invariant translation-direction error and the rotation error,
		\begingroup
		\setlength{\abovedisplayskip}{3pt}
		\setlength{\belowdisplayskip}{3pt}
		\setlength{\abovedisplayshortskip}{3pt}
		\setlength{\belowdisplayshortskip}{3pt}
		\begin{equation*}\label{eqn::pose_error}
			\max\left[\arccos\!\left(\frac{|\hat{\bft}^{\top}\bft^o|}{\|\hat{\bft}\|\,\|\bft^o\|}\right), \arccos\!\left(\frac{1}{2}{\rm tr}(\hat{\bfR}^\top\bfR^o)-\frac{1}{2}\right)\right],
		\end{equation*}
		\endgroup
		where hats and the superscript $o$ denote estimated and ground-truth quantities. A failed estimate receives an error of $180^\circ$. At threshold $\tau$, we integrate the linearly interpolated empirical recall curve $\widehat F$ using the trapezoidal convention to obtain the normalized \textit{Pose AUC}@$\tau$:
		\begingroup
		\setlength{\abovedisplayskip}{3pt}
		\setlength{\belowdisplayskip}{3pt}
		\setlength{\abovedisplayshortskip}{3pt}
		\setlength{\belowdisplayshortskip}{3pt}
		\begin{equation*}\label{eqn::pose_auc}
			\mathrm{AUC}@\tau=\frac{100}{\tau}\int_0^\tau \widehat F(e)\,\mathrm{d}e,
			\qquad \tau\in\{5^\circ,10^\circ,20^\circ\}.
		\end{equation*}
		\endgroup
		\textbf{Common practical configuration.} For RGB inputs, we resize the longer image edge to 1024 pixels; thermal inputs remain $640\times512$. SuperPoint, LightGlue, OmniGlue, and our method use the same SuperPoint detections~(NMS radius~4, $N_{\rm kp}=2048$). General-purpose descriptors are bilinearly sampled at SuperPoint detections and $\ell_2$-normalized. The m-to-m front end expands MKNN up to $K=5$ subject to $N_{\rm a}=2048$, applies our m-to-m adaptation of GMS~(threshold~4, grid~20, five scales/eight rotations), and retains at most $N_{\rm e}=1024$ associations. We use $q=0.3$, $\delta=0.01$, and a 1-pixel threshold. Adaptive runs use $T_{\min}/T_{\max}=10^3/10^5$, $\eta=0.9999$, and $\kappa=3$; our two-stage LO-RANSAC uses a pool capacity of $N_c=100$.

		\textbf{M-to-M estimator details.} The sampler makes $A_{\max}=\max(50,10n_x^+)$ attempts per draw, where $n_x^+$ counts active source vertices; the two-stage pool keeps the top $N_c$ raw seeds. After unprojection, an edge is an inlier if $r_{\rm S}^2<\epsilon^2$ and cheirality holds with a tolerance of $0.01$, where $\epsilon=\tfrac12(f_1^{-1}+f_2^{-1})\epsilon_{\rm px}$, $\epsilon_{\rm px}=1$, and $f_1,f_2$ are focal lengths. \textsc{LocalRefine} gathers edges with $r_{\rm S}^2<5\epsilon^2$~(i.e., $\tilde\epsilon=\sqrt{5}\epsilon$), then runs up to 25 PoseLib iterations using truncated least squares, $\rho(u)=\min(u,\epsilon^2)$, where $u$ is the squared refinement residual. \textsc{FinalPolish} reselects edges with $r_{\rm S}^2<\epsilon^2$ and runs up to 100 iterations using a Cauchy loss with scale~$\epsilon$. Both stages weight edge $(i,j)$ by $1/d_i+1/d_j$ using current inlier degrees.
		
		\textbf{Q1 fixed-proposal protocol.} Q1 retains only the $N_{\rm kp}=2048$ keypoint cap; it does not use GMS or the $N_{\rm a}/N_{\rm e}$ caps. It samples $E_1$ for all $10^5$ iterations and uses uncapped $E_K$ for assignment, scoring, refinement, and re-ranking. The ground-truth diagnostic is read-only and does not affect estimation.

		\textbf{Q2 estimator comparison.} All configurations share the ordered post-GMS graph~(first $N_{\rm e}=1024$ edges) and five pair-specific seeds. HCM-only and MCM-only use Algorithm~\ref{alg::m2m_loransac}; HCM$\rightarrow$MCM uses Algorithm~\ref{alg::m2m_RANSAC}. PoseLib CM treats the same edges as independent and unweighted while retaining its native sampling, scoring, and LO, thereby providing a conventional baseline with the same input.

		\textbf{Q3 representations and protocol.} We evaluate corrected layer-19 DINOv3 ViT-L/16~($s=200$)~\cite{simeoni2025dinov3,cuttano2026insid3}, raw final-layer DINOv2 ViT-L/14-register~\cite{oquab2023dinov2}, final-layer V-JEPA~2.1 ViT-L/16~\cite{murlabadia2026vjepa2_1}, final-layer SigLIP~2 So400m/16 NaFlex~\cite{tschannen2025siglip2}, and native SuperPoint as the task-specific control. V-JEPA~2.1 and SigLIP~2 have parameter counts comparable to those of the DINO models but use different pretraining objectives. For SigLIP~2, we choose NaFlex to accommodate diverse aspect ratios. Unlike the other representations, for which the longer RGB image edge is resized to 1,024 pixels and thermal images retain their original size, SigLIP~2's officially recommended 1,024-patch budget induces an aspect-ratio-preserving, input-dependent resize; valid tokens are retained and mapped to image coordinates with half-pixel scaling. All sampled descriptors are $\ell_2$-normalized. Within each representation, MNN+PoseLib LO-RANSAC is compared with the practical m-to-m pipeline above. They share detections, pairs, threshold, $N_{\rm e}$, five seeds, and downstream settings; only the m-to-m pipeline uses the pre-GMS $N_{\rm a}$ cap.

		\begin{table*}[!htbp]
			\centering
			\caption{Mean \textit{Pose AUC}@$5^\circ\!/10^\circ\!/20^\circ$~(\%) over five repetitions for the same-graph Q2 comparison. \textbf{Bold} indicates the highest unrounded mean in each cell.}
			\label{table::q2_estimator_auc}
			\small
			\resizebox{\textwidth}{!}{%
			\begin{tabular}{c|cccccc}
				\hline
				Configuration & ScanNet-1500 & MegaDepth-1500 & NAVI-Multi & NAVI-Wild & METU-CC & METU-CS\\
				\hline
				m-to-m + PoseLib CM & 11.6 / 24.1 / 38.5 & 22.1 / 33.3 / 46.8 & 10.7 / 21.9 / 32.7 & 5.5 / 13.9 / 24.3 & 11.0 / 23.2 / 37.8 & 4.5 / 14.6 / 31.3\\
				HCM & 13.2 / 28.0 / 44.1 & 28.2 / 40.7 / 54.4 & 11.0 / 23.1 / 34.6 & 6.0 / 14.9 / 26.4 & 13.0 / 26.4 / 42.5 & 5.3 / 17.3 / 35.5\\
				MCM & 13.8 / 29.3 / 46.0 & 33.1 / 47.0 / 60.5 & 11.1 / 23.0 / 34.6 & 5.9 / 14.9 / 26.5 & 13.4 / 27.7 / 44.6 & \textbf{5.7} / \textbf{18.3} / 37.4\\
				HCM$\rightarrow$MCM & \textbf{13.9} / \textbf{29.4} / \textbf{46.4} & \textbf{34.0} / \textbf{47.5} / \textbf{60.8} & \textbf{11.6} / \textbf{23.9} / \textbf{35.4} & \textbf{6.4} / \textbf{15.5} / \textbf{27.1} & \textbf{13.7} / \textbf{28.2} / \textbf{45.1} & 5.6 / 18.3 / \textbf{37.5}\\
				\hline
			\end{tabular}}
		\end{table*}
	\section{Additional Experimental Analyses}
	\label{append::ablation_and_sensitivity}
	\subsection{Q1: Fixed-Proposal \texorpdfstring{$K$}{K}-Sweep Diagnostics}
	\label{append::fixed_k_diagnostics}
	For each pair--repetition combination and each $K$, we score the reference pose with HCM and ask whether it would strictly enter the top-100 raw Stage-1 pool. Table~\ref{table::gt_hypothesis_inclusion} reports the resulting rates. 
	\begin{table}[!htb]
		\small
		\setlength{\tabcolsep}{5pt}
		\centering
		\caption{Reference-pose top-100 entry rate~(\%), pooled over pair--seed trials. Macro denotes the average across datasets; ties at rank 100 are excluded.}
		\label{table::gt_hypothesis_inclusion}
		\begin{tabular}{l|rrrrr}
			\hline
			Dataset & $K=1$ & $K=2$ & $K=3$ & $K=4$ & $K=5$\\
			\hline
			ScanNet & 1.60 & 3.61 & 5.43 & 5.61 & 5.65\\
			MegaDepth & 55.06 & 55.87 & 52.05 & 47.42 & 42.49\\
			NAVI-Multi & 7.33 & 14.57 & 18.85 & 19.85 & 22.08\\
			NAVI-Wild & 2.71 & 7.49 & 10.84 & 12.63 & 14.37\\
			METU-CC & 0.51 & 2.08 & 3.31 & 4.11 & 4.37\\
			METU-CS & 0.17 & 2.17 & 3.54 & 4.32 & 5.23\\
			\hline
			Macro & 11.23 & 14.30 & 15.67 & 15.66 & 15.70\\
			\hline
		\end{tabular}
	\end{table}
	\subsection{Q2: Estimator Comparison}
	\label{append::q2_estimator_scores}
	\begin{figure}[!htb]
		\centering
		\includegraphics[width=\linewidth]{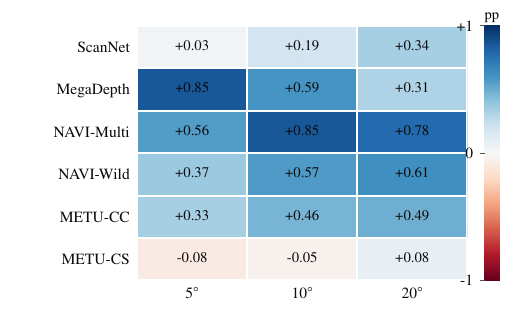}
		\caption{Gain of the two-stage estimator over the better single-stage estimator, averaged over five matched seeds. Each cell reports, in pp, the mean \textit{Pose AUC} of HCM$\rightarrow$MCM minus the larger of the corresponding HCM-only and MCM-only means. Positive values in 16 of 18 cells indicate complementarity of these two robust mechanisms; the METU-CS exceptions at $5^\circ$ and $10^\circ$ are within $0.1$~pp.}
		\label{fig::q2_complementarity}
	\end{figure}
	\begin{figure*}[!b]
		\centering
		\includegraphics[width=0.88\textwidth]{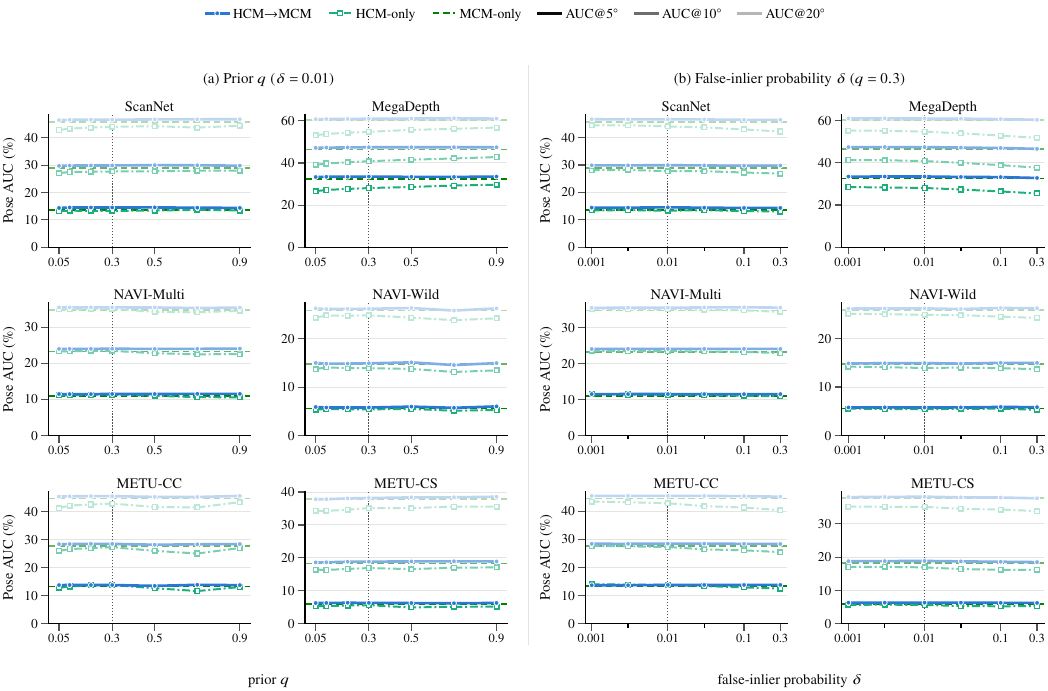}
		\caption{Single-repetition sensitivity under the practical configuration. \textbf{(a)}~$q$ at $\delta=0.01$; values with $q>0.3$ are included as score-scale stress tests. \textbf{(b)}~$\delta$ at $q=0.3$~(logarithmic axis). Dots mark the defaults; curves show \textit{Pose AUC}@$5^\circ/10^\circ/20^\circ$.}
		\label{fig::hcm_sensitivity}
	\end{figure*}
	Table~\ref{table::q2_estimator_auc} reports the mean \textit{Pose AUC} over five pair-specific seeds on shared post-GMS association graphs. Figure~\ref{fig::q2_complementarity} reports the mean \textit{Pose AUC} gain of HCM$\rightarrow$MCM over the better of HCM-only and MCM-only for each dataset--threshold pair. Across the four configurations and all dataset--threshold pairs, the largest sample standard deviation over seeds is $0.86$~pp.
	
	\subsection{Q3: Representation Compatibility}
	\label{append::variant_scores}
	Within each representation, Q3 compares MNN+PoseLib CM with the fixed preserve-then-resolve pipeline~(progressive MKNN with $K=5$, followed by HCM$\rightarrow$MCM). The four general-purpose representations match the main text: corrected layer-19 DINOv3 ViT-L/16, raw final-layer DINOv2 ViT-L/14-register, final-layer V-JEPA~2.1 ViT-L/16, and final-layer SigLIP~2 So400m/16 NaFlex. The latter two test whether the ambiguity mechanism extends beyond DINO-style visual representations. Native SuperPoint descriptors provide a task-specific control, distinct from the conventional SuperPoint system in Table~\ref{table::relative_pose_accuracy}.

	Table~\ref{table::variant_auc} reports absolute accuracy, and Table~\ref{table::q3_dataset_delta} gives the paired gains. For each repetition and each NAVI dataset, we pool the three 500-pair viewpoint bins before computing each \textit{Pose AUC}; the dataset average then weights the six test sets equally. Paired $\Delta\mathcal{A}$ subtracts MNN+PoseLib CM from preserve-then-resolve under the same random seed. 
	\begin{table*}[!p]
		\centering
		\caption{Within-representation Q3 mean \textit{Pose AUC}@$5^\circ\!/10^\circ\!/20^\circ$~(\%) over five pair-specific random seeds~(nearest $0.1$~pp). MNN+CM denotes MNN+PoseLib CM; m-to-m+H$\rightarrow$M denotes preserve-then-resolve.}
		\label{table::variant_auc}
		\scriptsize
		\setlength{\tabcolsep}{2.5pt}
		\begin{tabular}{ll|cccccc}
			\hline
			Representation & Pipeline & ScanNet-1500 & MegaDepth-1500 & NAVI-Multi & NAVI-Wild & METU-CC & METU-CS\\
			\hline
			\multirow{2}{*}{Corrected DINOv3} & MNN+CM & 10.8 / 23.8 / 39.7 & 32.4 / 46.3 / 60.0 & 9.7 / 20.3 / 31.0 & 4.0 / 10.4 / 19.8 & 7.8 / 18.7 / 34.3 & 4.0 / 13.6 / 29.9\\
			 & m-to-m+H$\rightarrow$M & 13.9 / 29.4 / 46.4 & 34.0 / 47.5 / 60.8 & 11.6 / 23.9 / 35.4 & 6.4 / 15.5 / 27.1 & 13.7 / 28.2 / 45.1 & 5.6 / 18.3 / 37.5\\
			\hline
			\multirow{2}{*}{Raw DINOv2} & MNN+CM & 3.6 / 8.7 / 16.8 & 18.2 / 29.1 / 42.2 & 4.1 / 9.7 / 16.8 & 2.0 / 5.5 / 11.9 & 2.8 / 9.2 / 20.8 & 1.7 / 7.3 / 19.7\\
			 & m-to-m+H$\rightarrow$M & 6.2 / 14.3 / 25.6 & 19.8 / 30.9 / 44.9 & 5.7 / 13.2 / 22.3 & 3.5 / 9.7 / 19.4 & 5.4 / 14.5 / 28.0 & 3.0 / 11.1 / 26.0\\
			\hline
			\multirow{2}{*}{V-JEPA~2.1} & MNN+CM & 8.1 / 18.3 / 31.5 & 26.6 / 39.0 / 51.7 & 6.7 / 14.2 / 21.9 & 1.8 / 5.7 / 11.8 & 6.2 / 15.6 / 29.3 & 2.3 / 9.7 / 24.3\\
			 & m-to-m+H$\rightarrow$M & 10.9 / 23.7 / 38.8 & 28.8 / 41.5 / 54.0 & 8.6 / 17.8 / 26.7 & 3.6 / 9.2 / 17.6 & 9.2 / 20.9 / 36.4 & 3.5 / 13.6 / 31.0\\
			\hline
			\multirow{2}{*}{SigLIP~2 NaFlex} & MNN+CM & 0.6 / 2.2 / 5.9 & 2.7 / 6.2 / 12.8 & 0.4 / 1.4 / 3.9 & 0.2 / 0.6 / 2.1 & 0.5 / 2.1 / 7.0 & 0.1 / 0.8 / 4.3\\
			 & m-to-m+H$\rightarrow$M & 2.0 / 6.2 / 13.8 & 7.4 / 14.4 / 24.7 & 1.5 / 4.7 / 10.3 & 0.7 / 2.4 / 6.6 & 1.1 / 3.8 / 10.8 & 0.4 / 1.6 / 6.6\\
			\hline
			\multirow{2}{*}{SuperPoint descriptor} & MNN+CM & 13.7 / 27.6 / 42.0 & 44.0 / 58.1 / 68.9 & 11.5 / 22.2 / 31.4 & 2.6 / 7.0 / 12.6 & 0.4 / 1.2 / 3.3 & 0.7 / 1.9 / 4.5\\
			 & m-to-m+H$\rightarrow$M & 13.4 / 27.5 / 42.5 & 44.9 / 58.9 / 69.9 & 12.1 / 23.3 / 32.4 & 3.2 / 8.1 / 14.6 & 0.6 / 1.6 / 3.9 & 1.1 / 2.9 / 6.5\\
			\hline
		\end{tabular}
		\vspace{4pt}

		\centering
		\caption{Paired Q3 $\Delta\mathcal{A}$~(preserve-then-resolve minus MNN+PoseLib CM, pp). Entries report the five-seed mean $\pm$ sample standard deviation~[min, max]~(nearest $0.01$~pp); ``Dataset avg.'' weights the six test sets equally.}
		\label{table::q3_dataset_delta}
		\footnotesize
		\renewcommand{\arraystretch}{1.35}
		\setlength{\tabcolsep}{3.0pt}
		\begin{tabular}{ll|ccc}
			\hline
			Representation & Dataset & $\Delta\mathcal{A}$@$5^\circ$ & $\Delta\mathcal{A}$@$10^\circ$ & $\Delta\mathcal{A}$@$20^\circ$\\
			\hline
			\multirow{7}{*}{Corrected DINOv3}
			 & ScanNet-1500 & $3.10\!\pm\!0.28\,[2.82,3.49]$ & $5.60\!\pm\!0.36\,[5.14,6.00]$ & $6.70\!\pm\!0.46\,[5.97,7.21]$\\
			 & MegaDepth-1500 & $1.53\!\pm\!0.76\,[0.36,2.15]$ & $1.29\!\pm\!0.67\,[0.17,1.76]$ & $0.78\!\pm\!0.39\,[0.13,1.11]$\\
			 & NAVI-Multi & $1.97\!\pm\!0.13\,[1.79,2.12]$ & $3.60\!\pm\!0.34\,[3.21,4.13]$ & $4.41\!\pm\!0.44\,[3.91,5.05]$\\
			 & NAVI-Wild & $2.36\!\pm\!0.22\,[2.05,2.64]$ & $5.16\!\pm\!0.23\,[4.82,5.38]$ & $7.26\!\pm\!0.21\,[7.06,7.57]$\\
			 & METU-CC & $5.95\!\pm\!0.37\,[5.58,6.53]$ & $9.47\!\pm\!0.51\,[9.01,10.30]$ & $10.78\!\pm\!0.77\,[10.03,11.97]$\\
			 & METU-CS & $1.62\!\pm\!0.54\,[0.98,2.49]$ & $4.66\!\pm\!0.62\,[3.97,5.52]$ & $7.58\!\pm\!0.44\,[7.12,8.24]$\\
			 & Dataset avg. & $2.75\!\pm\!0.14\,[2.58,2.90]$ & $4.96\!\pm\!0.19\,[4.72,5.25]$ & $6.25\!\pm\!0.20\,[5.97,6.42]$\\
			\hline
			\multirow{7}{*}{Raw DINOv2}
			 & ScanNet-1500 & $2.65\!\pm\!0.22\,[2.34,2.91]$ & $5.54\!\pm\!0.34\,[4.96,5.79]$ & $8.80\!\pm\!0.57\,[8.11,9.59]$\\
			 & MegaDepth-1500 & $1.58\!\pm\!0.43\,[1.01,2.09]$ & $1.88\!\pm\!0.68\,[0.81,2.57]$ & $2.70\!\pm\!0.74\,[1.50,3.41]$\\
			 & NAVI-Multi & $1.65\!\pm\!0.16\,[1.38,1.80]$ & $3.50\!\pm\!0.42\,[3.01,3.95]$ & $5.47\!\pm\!0.64\,[4.88,6.51]$\\
			 & NAVI-Wild & $1.48\!\pm\!0.31\,[1.14,1.79]$ & $4.20\!\pm\!0.44\,[3.72,4.83]$ & $7.42\!\pm\!0.53\,[6.69,8.02]$\\
			 & METU-CC & $2.60\!\pm\!0.42\,[2.11,3.26]$ & $5.27\!\pm\!0.33\,[4.85,5.73]$ & $7.17\!\pm\!0.57\,[6.33,7.77]$\\
			 & METU-CS & $1.30\!\pm\!0.22\,[1.07,1.66]$ & $3.80\!\pm\!0.47\,[3.47,4.61]$ & $6.27\!\pm\!0.87\,[5.40,7.58]$\\
			 & Dataset avg. & $1.88\!\pm\!0.12\,[1.76,2.07]$ & $4.03\!\pm\!0.19\,[3.86,4.28]$ & $6.31\!\pm\!0.27\,[5.94,6.60]$\\
			\hline
			\multirow{7}{*}{V-JEPA~2.1}
			 & ScanNet-1500 & $2.74\!\pm\!0.50\,[1.87,3.10]$ & $5.40\!\pm\!0.63\,[4.35,5.89]$ & $7.34\!\pm\!0.44\,[6.90,7.82]$\\
			 & MegaDepth-1500 & $2.22\!\pm\!0.23\,[1.90,2.46]$ & $2.48\!\pm\!0.53\,[1.83,3.26]$ & $2.22\!\pm\!0.53\,[1.69,3.06]$\\
			 & NAVI-Multi & $1.89\!\pm\!0.46\,[1.36,2.46]$ & $3.54\!\pm\!0.27\,[3.21,3.81]$ & $4.78\!\pm\!0.51\,[4.08,5.47]$\\
			 & NAVI-Wild & $1.75\!\pm\!0.25\,[1.47,2.01]$ & $3.55\!\pm\!0.34\,[3.28,4.12]$ & $5.84\!\pm\!0.40\,[5.24,6.32]$\\
			 & METU-CC & $3.00\!\pm\!0.68\,[1.85,3.59]$ & $5.33\!\pm\!0.91\,[3.78,6.12]$ & $7.09\!\pm\!0.77\,[5.75,7.71]$\\
			 & METU-CS & $1.22\!\pm\!0.24\,[0.91,1.59]$ & $3.90\!\pm\!0.42\,[3.37,4.46]$ & $6.64\!\pm\!0.53\,[5.83,7.27]$\\
			 & Dataset avg. & $2.14\!\pm\!0.26\,[1.71,2.42]$ & $4.03\!\pm\!0.31\,[3.51,4.26]$ & $5.65\!\pm\!0.23\,[5.26,5.84]$\\
			\hline
			\multirow{7}{*}{SigLIP~2 NaFlex}
			 & ScanNet-1500 & $1.45\!\pm\!0.35\,[0.93,1.79]$ & $4.05\!\pm\!0.28\,[3.64,4.32]$ & $7.93\!\pm\!0.27\,[7.66,8.34]$\\
			 & MegaDepth-1500 & $4.78\!\pm\!0.18\,[4.49,4.95]$ & $8.19\!\pm\!0.38\,[7.74,8.81]$ & $11.87\!\pm\!0.43\,[11.51,12.53]$\\
			 & NAVI-Multi & $1.08\!\pm\!0.17\,[0.85,1.24]$ & $3.26\!\pm\!0.26\,[2.96,3.67]$ & $6.35\!\pm\!0.27\,[5.87,6.53]$\\
			 & NAVI-Wild & $0.52\!\pm\!0.09\,[0.39,0.60]$ & $1.83\!\pm\!0.18\,[1.63,2.01]$ & $4.51\!\pm\!0.27\,[4.36,4.99]$\\
			 & METU-CC & $0.60\!\pm\!0.28\,[0.32,1.00]$ & $1.76\!\pm\!0.30\,[1.36,2.12]$ & $3.80\!\pm\!0.40\,[3.36,4.29]$\\
			 & METU-CS & $0.23\!\pm\!0.11\,[0.06,0.33]$ & $0.81\!\pm\!0.27\,[0.35,0.99]$ & $2.23\!\pm\!0.36\,[1.86,2.71]$\\
			 & Dataset avg. & $1.44\!\pm\!0.12\,[1.31,1.54]$ & $3.32\!\pm\!0.08\,[3.23,3.44]$ & $6.11\!\pm\!0.13\,[5.95,6.30]$\\
			\hline
			\multirow{7}{*}{SuperPoint descriptor}
			 & ScanNet-1500 & $-0.30\!\pm\!0.54\,[-0.97,0.45]$ & $-0.09\!\pm\!0.81\,[-1.23,0.88]$ & $0.50\!\pm\!0.82\,[-0.77,1.32]$\\
			 & MegaDepth-1500 & $0.82\!\pm\!0.37\,[0.38,1.28]$ & $0.88\!\pm\!0.17\,[0.70,1.15]$ & $1.02\!\pm\!0.15\,[0.90,1.27]$\\
			 & NAVI-Multi & $0.55\!\pm\!0.33\,[0.05,0.90]$ & $1.13\!\pm\!0.34\,[0.59,1.43]$ & $1.04\!\pm\!0.45\,[0.38,1.48]$\\
			 & NAVI-Wild & $0.57\!\pm\!0.24\,[0.23,0.88]$ & $1.11\!\pm\!0.36\,[0.53,1.42]$ & $2.04\!\pm\!0.52\,[1.39,2.75]$\\
			 & METU-CC & $0.21\!\pm\!0.10\,[0.08,0.33]$ & $0.39\!\pm\!0.17\,[0.24,0.65]$ & $0.61\!\pm\!0.24\,[0.40,0.97]$\\
			 & METU-CS & $0.36\!\pm\!0.20\,[0.17,0.63]$ & $0.97\!\pm\!0.27\,[0.68,1.41]$ & $1.95\!\pm\!0.35\,[1.57,2.39]$\\
			 & Dataset avg. & $0.37\!\pm\!0.10\,[0.21,0.48]$ & $0.73\!\pm\!0.16\,[0.47,0.92]$ & $1.19\!\pm\!0.21\,[0.90,1.41]$\\
			\hline
		\end{tabular}
	\end{table*}
	\subsection{HCM Hyperparameter Sensitivity}
	\label{section::sensitivity}
	Figure~\ref{fig::hcm_sensitivity} sweeps $q=0.05$--$0.9$ at $\delta=0.01$ and $\delta=0.001$--$0.3$ at $q=0.3$. For NAVI-Wild, we pool the three 500-pair viewpoint bins. For the $q>0.3$ stress-test points, the implementation applies a $10^{-9}$ floor to the endpoint denominators as a numerical safeguard. Because a complete grid is computationally expensive, this descriptive robustness check uses only repetition $b=0$, with pair-specific effective seeds $s_{i,0}=i$. The maximum peak-to-peak AUC variations of Algorithm~\ref{alg::m2m_RANSAC}~(HCM$\rightarrow$MCM) are $0.76$~pp over $q$ and $0.83$~pp over $\delta$. Across the 12 unique hyperparameter settings, counting the shared default only once, HCM$\rightarrow$MCM matches or exceeds MCM-only in 211 of 216 dataset--threshold cells; the largest deficit is $0.21$~pp.
\fi
\end{document}